\documentclass[10pt,journal,compsoc]{IEEEtran}
\ifCLASSOPTIONcompsoc
  \usepackage[nocompress]{cite}
\else
  \usepackage{cite}
\fi

\ifCLASSINFOpdf
\else
\fi

\usepackage[dvipsnames]{xcolor}
\usepackage{amsmath,amsfonts}
\usepackage{algorithmic}
\usepackage{algorithm}
\usepackage{array}
\usepackage{textcomp}
\usepackage{stfloats}
\usepackage{url}
\usepackage{verbatim}
\usepackage{graphicx}
\usepackage{cite}
\usepackage{enumitem}
\usepackage{subcaption}
\usepackage{colortbl}
\usepackage{multirow}
\usepackage{booktabs}
\usepackage[switch]{lineno}
\usepackage{svg}
\usepackage{amssymb}

\usepackage[pagebackref=false,breaklinks=true,letterpaper=true,colorlinks,bookmarks=false]{hyperref}

\definecolor{myblue}{rgb}{.894,.937,.965}
\definecolor{mydark_blue}{rgb}{.68,.776,.906}
\definecolor{NO1}{rgb}{.749,.902,.808}
\definecolor{NO2}{rgb}{.999,.973,.773}
\definecolor{NO3}{rgb}{255,255,255}

\usepackage{array}

\begin{document}
\title{3D Gaussian Splatting in Robotics: A Survey}

\author{Siting Zhu, Guangming Wang, Xin Kong, Dezhi Kong, Hesheng Wang~\IEEEmembership{Senior Member,~IEEE}
\thanks{\quad This work was supported in part by the Natural Science Foundation of China under Grant 62225309, 62073222, U21A20480 and 62361166632. (Corresponding author: Hesheng Wang.)}
\thanks{\quad Siting Zhu is with the Department of Automation, Shanghai Jiao Tong University, Shanghai. (email: zhusiting@sjtu.edu.cn) Guangming Wang is with the University of Cambridge. Xin Kong is with Imperial College London. Dezhi Kong is with Shanghai Jiao Tong University, Shanghai.}
\thanks{\quad Hesheng Wang is with Department of Automation, Key Laboratory of
System Control and Information Processing of Ministry of Education, Key
Laboratory of Marine Intelligent Equipment and System of Ministry of
Education, Shanghai Jiao Tong University. (email: wanghesheng@sjtu.edu.cn)}
}

\IEEEtitleabstractindextext{
\begin{abstract}
Dense 3D representations of the environment have been a long-term goal in the robotics field. While previous Neural Radiance Fields (NeRF) representation have been prevalent for its implicit, coordinate-based model, the recent emergence of 3D Gaussian Splatting (3DGS) has demonstrated remarkable potential in its explicit radiance field representation. By leveraging 3D Gaussian primitives for explicit scene representation and enabling differentiable rendering, 3DGS has shown significant advantages over other radiance fields in real-time rendering and photo-realistic performance, which is beneficial for robotic applications. In this survey, we provide a comprehensive understanding of 3DGS in the field of robotics. We divide our discussion of the related works into two main categories: the application of 3DGS and the advancements in 3DGS techniques. In the application section, we explore how 3DGS has been utilized in various robotics tasks from scene understanding and interaction perspectives. The advance of 3DGS section focuses on the improvements of 3DGS own properties in its adaptability and efficiency, aiming to enhance its performance in robotics. We then summarize the most commonly used datasets and evaluation metrics in robotics.
Finally, we identify the challenges and limitations of current 3DGS methods and discuss the future development of 3DGS in robotics. The paper lists of our survey are available at \href{https://github.com/zstsandy/Awesome-3D-Gaussian-Splatting-in-Robotics}{3D Gaussian Splatting in Robotics}.
\end{abstract}

\begin{IEEEkeywords}
3D Gaussian Splatting, Robotics, Scene Understanding and Interaction, Challenges and Future Directions
\end{IEEEkeywords}
}
\maketitle

\IEEEdisplaynontitleabstractindextext
\IEEEpeerreviewmaketitle

\ifCLASSOPTIONcompsoc
\IEEEraisesectionheading{\section{Introduction}\label{sec:introduction}}
\else
\section{Introduction}
\label{sec:introduction}
\fi

\IEEEPARstart{T}he advent of Neural Radiance Fields (NeRF)~\cite{mildenhall2021nerf} has promoted the development of robotics, particularly enhancing robots' capabilities in perception, scene reconstruction and interaction with their environments. However, this implicit representation suffers from its inefficient per-pixel raycasting rendering for optimization. The emergence of 3D Gaussian Splatting (3DGS)~\cite{kerbl20233d} addresses this inefficiency by its explicit representation and achieves high-quality and real-time rendering through splatting. Specifically, 3DGS models the environment using a set of Gaussian primitives with learnable parameters, providing explicit representation of scenes. In rendering process, 3DGS employs splatting~\cite{westover1991splatting} to project 3D Gaussians into 2D image space given camera poses and applies tile-based rasterizer for acceleration, enabling real-time performance. Therefore, 3DGS has greater potential to promote the performance and expand the capabilities of robotic systems.

With the debut of 3DGS in 2023, several survey papers~\cite{chen2024survey, fei20243d, dalal2024gaussian, wu2024recent, bagdasarian20243dgs, bao20243d} have been published to show the developments in this area. Chen {\it et al.}~\cite{chen2024survey} presented the first survey of 3DGS. This survey describes the recent developments and critical contributions of 3DGS methods. Fei {\it et al.}~\cite{fei20243d} introduced a unified framework for categorizing existing works in 3DGS. Wu {\it et al.}~\cite{wu2024recent} presented a survey including both traditional splatting methods and recent neural-based 3DGS methods, showing the development in splatting techniques of 3DGS. Bao {\it et al.}~\cite{bao20243d} provided a more detailed classification based on technologies of 3DGS. Additionally, Dalal {\it et al.}~\cite{dalal2024gaussian} focused on the tasks of 3D Reconstruction in 3DGS and Bagdasarian {\it et al.}~\cite{bagdasarian20243dgs} summarized 3DGS-based compression methods, which demonstrates the weaknesses and advantages of 3DGS in specific domains.

\begin{figure*}[!t]
\centering 
\includegraphics[width=\linewidth]{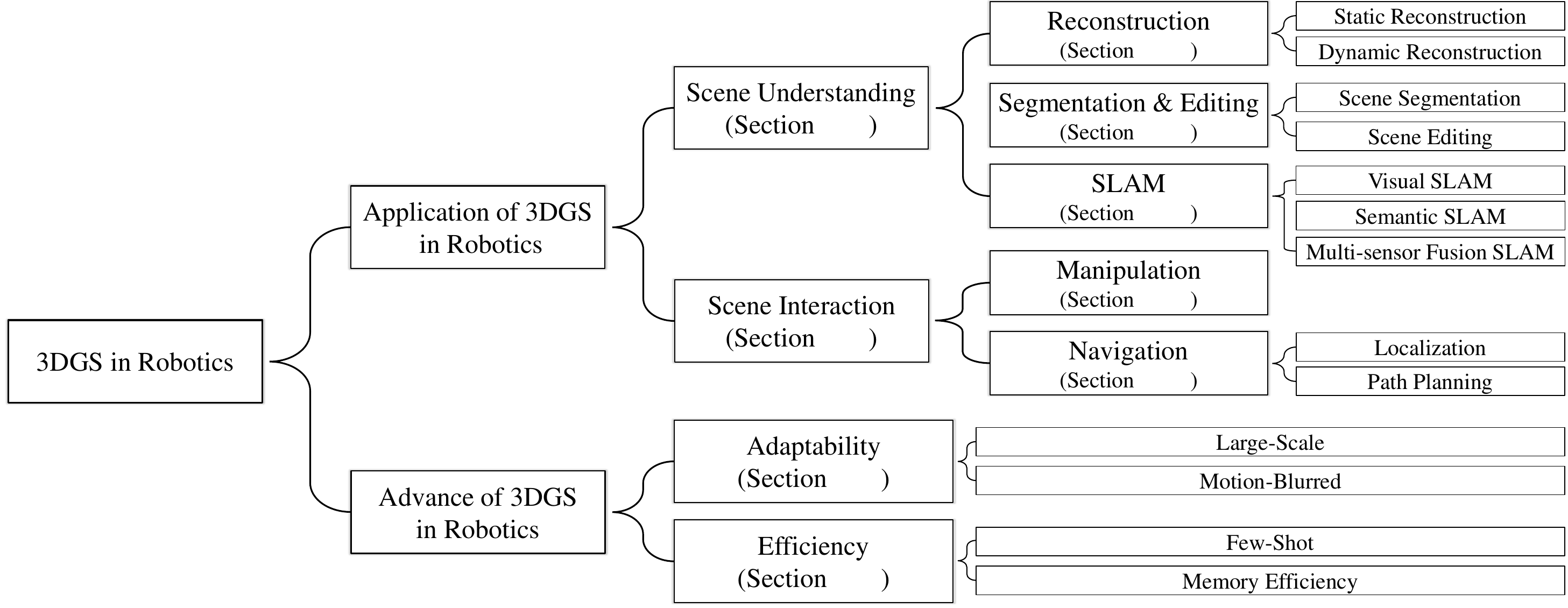}
\put(-140.3, 181.8){\scriptsize{\ref{sec:Reconstruction}}}
\put(-140.3, 154.8){\scriptsize{\ref{sec:Segmentation}}}
\put(-140.3, 127.8){\scriptsize{\ref{sec:SLAM}}}
\put(-140.3, 99.8){\scriptsize{\ref{sec:Manipulation}}}
\put(-140.3, 72.8){\scriptsize{\ref{sec:Navigation}}}
\put(-245.3, 156.3){\small{\ref{sec:Scene Understanding}}}
\put(-245.3, 85.7){\small{\ref{sec:Scene Interaction}}}
\put(-242, 39.2){\small{\ref{sec:Adaptability}}}
\put(-242, 6.2){\small{\ref{sec:Efficiency}}}
\vspace{-0.02in}
\caption{A taxonomy of 3D Gaussian Splatting (3DGS) in Robotics.}
\vspace{-0.05in}
\label{fig:overview}
\end{figure*}

However, existing 3DGS surveys either provide broad categorizations of 3DGS works or focus on real-time view synthesis of 3DGS, which lacks detailed summaries in the field of robotics. Therefore, in this paper, we provide a comprehensive summary and detailed classification of 3DGS in robotics. We introduce application of 3DGS in robotics and provide a detailed classification of related works regarding 3DGS-based robotic applications. Moreover, we summarize the potential solutions to enhance 3DGS representation for robotic systems. Finally, we show performance evaluation of 3DGS-based works and discuss the future development of 3DGS in robotics. The overall framework of our survey is demonstrated in Fig.~\ref{fig:overview}.

Section~\ref{sec:background} provides a brief background on core concepts and mathematical principles of 3DGS.
Section~\ref{sec:application} categorizes various application directions of 3DGS in robotics and presents a detailed classification of related works in specific directions. 
Section \ref{sec:advance} discusses various advances that improve the representation of 3DGS, aiming to enhance its capability for robotics tasks.
Additionally, in Section~\ref{sec:performance}, we summarize the datasets and evaluation metrics that are used in applications of 3DGS in robotics. Moreover, this section also compares the performance of existing methods in different robotic directions.
In Section~\ref{sec:future}, we discuss the challenges and future directions of 3DGS in robotics.
Finally, Section~\ref{sec:conclusion} presents the conclusion of this survey.

\begin{figure*}[!t]
\centering 
\includegraphics[width=\linewidth]{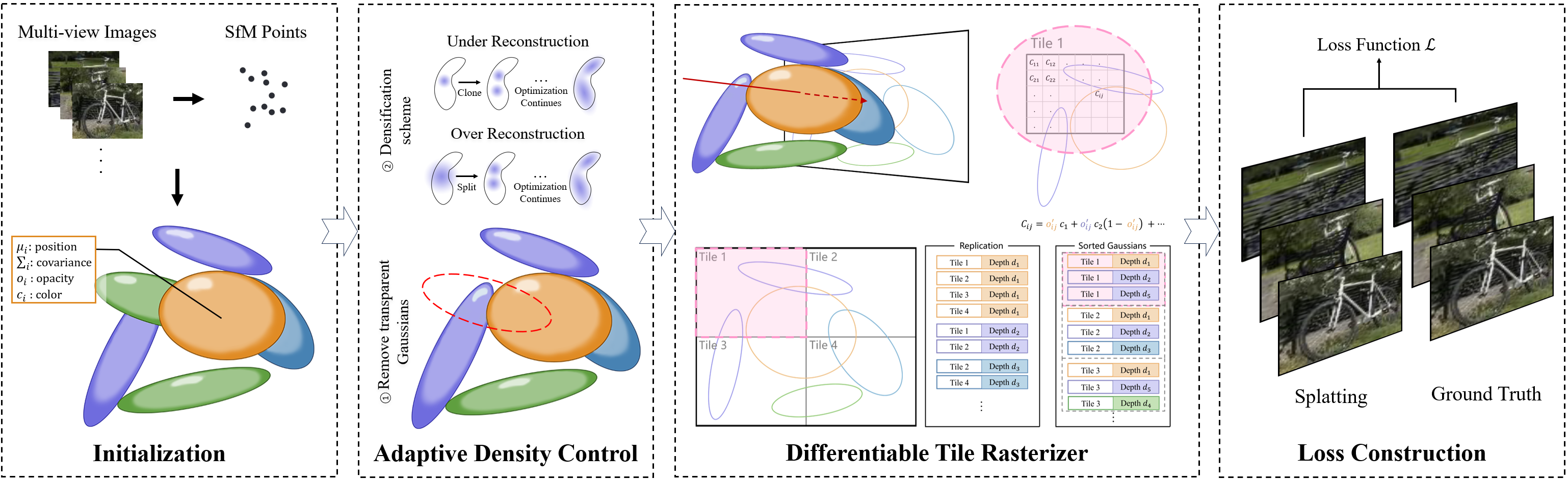}
\vspace{-0.15in}
\caption{An illustration of the forward process of 3DGS.}
\vspace{-0.1in}
\label{fig:3dgs}
\end{figure*}

\section{Background}
\label{sec:background}
\subsection{3DGS Theory}
3DGS~\cite{kerbl20233d} introduces an explicit radiance field representation for real-time and high-quality rendering of 3D scenes. This technique models the environment using a set of 3D Gaussian primitives, denoted as $G = \{g_1, g_2, \ldots, g_n\}$. Each 3D Gaussian $g_i$ is parameterized by a set of learnable properties, including the 3D center position $\mu_i \in \mathbb{R}^3$, 3D covariance matrix $\Sigma_i \in \mathbb{R}^{3\times 3}$, opacity $o_i \in [0,1]$, and color $c_i$ represented by spherical harmonics (SH) coefficients ~\cite{cabral1987bidirectional} for view-dependent appearance. These properties allow for the spatial compact representation of an individual Gaussian as:
\begin{equation}
g_i(x) = e^{-\frac{1}{2}(x-\mu_i)^T \Sigma_i^{-1} (x-\mu_i)}.
\end{equation}
Here, covariance matrix $\Sigma$ of 3D Gaussian is analogous to describing the configuration of an ellipsoid, and is computed as \(\Sigma=R S S^T R^T\). $S$ is a scaling matrix and $R$ represents the rotation. 

3DGS takes multi-view images as input, generating a sparse point cloud through Structure-from-Motion (SfM)~\cite{schonberger2016structure}, which is then used to initialize 3D Gaussian primitives. The 3D Gaussian representation is subsequently optimized by adjusting the properties of these primitives to minimize the difference between rendered and ground-truth images.
Rendering results in 3DGS are produced using a splatting process alongside a differentiable, tile-based rasterizer designed for acceleration. Additionally, 3DGS incorporates adaptive density control to optimize the number of 3D Gaussians used in scene representation. Fig.~\ref{fig:3dgs} illustrates the forward process of 3DGS. The following sections detail the splatting process, which enables fast rendering, the loss function employed in 3D Gaussian optimization, and the adaptive density control technique. This adaptive control allows for progressively densifying an initially sparse Gaussian set, resulting in a more refined scene representation.

\noindent\textbf{Splatting.}\hspace{5pt}
In this process, 3D Gaussians are projected to 2D image space for rendering. 
Given the viewing transformation $W$, the projected 2D covariance matrix in camera coordinates is computed as  $\Sigma' = JW \Sigma W^T J^T$, where $J$ is the Jacobian of the affine approximation of the projective transformation. Therefore, the final pixel color $C$ can be computed by $\alpha$-blending 3D Gaussian splats that overlap at a given pixel, with the Gaussians sorted in depth order:
\begin{equation}
C = \sum_{i \in N} c_i \alpha_i \prod_{j=1}^{i-1} (1 - \alpha_j),
\end{equation}
where the final opacity $\alpha_i$ is formulated as:
\begin{equation}
\alpha_i = o_i \exp\left(-\frac{1}{2}(x' - \mu'_i)^T \Sigma'^{-1}_i (x' - \mu'_i)\right),
\end{equation}
where $x'$ and $\mu_i'$ are coordinates in the projected space. 

\noindent\textbf{Loss Function.}\hspace{5pt}
For the optimization of 3D Gaussian properties, the loss is constructed as the difference between splatting image and ground truth image. Specifically, the loss function is a combination of $\mathcal{L}_1$ loss and a D-SSIM term:
\begin{equation}
\mathcal{L} = (1 - \lambda)\mathcal{L}_1 + \lambda \mathcal{L}_{\text{D-SSIM}},
\end{equation}
where $\lambda$ is a weighting coefficient and is set to 0.2 in 3DGS.

\noindent\textbf{Adaptive Density Control.}\hspace{5pt}
In the process of Gaussian optimization, 3DGS employs periodic adaptive densification for detailed reconstruction. This strategy focuses on areas with missing geometric features or regions where Gaussians are over-expanded, both exhibiting large view-space positional gradients. For under-reconstructed areas, small Gaussians are cloned and moved towards the positional gradient direction. In over-reconstructed regions, large Gaussians with high variance are split into two smaller ones. Additionally, 3DGS removes Gaussians that are virtually transparent, with opacity less than a specific threshold.

In conclusion, 3DGS introduces a novel explicit radiance field representation using 3D Gaussian primitives, which offers a compact, efficient, and flexible approach to model 3D scenes. The use of 3D Gaussian representation, combined with the splatting process, differentiable optimization, and adaptive density control, makes 3DGS a powerful tool for real-time and high-quality rendering of complex 3D environments.
\section{Application of 3DGS in Robotics}
\label{sec:application}
The advantages of 3DGS, including its explicit radiance field representation and fast rendering capability, make it an attractive representation option for robotics applications. These properties are crucial for achieving comprehensive scene understanding in robotics and executing particular tasks through interaction with the environment.

\begin{figure*}[!t]
\centering 
\includegraphics[width=\linewidth]{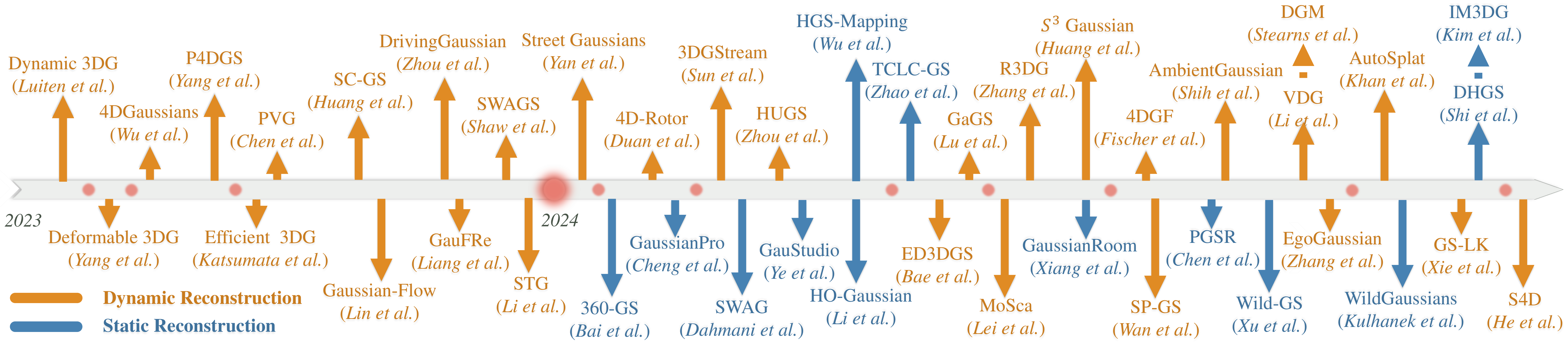}
\vspace{-0.2in}
\caption{Chronological: 3DGS for Scene Reconstruction. The red dots in the figure represent months, which also applys to the other figures.}
\vspace{-0.1in}
\label{fig:recon_time}
\end{figure*}
\subsection{Scene Understanding}
\label{sec:Scene Understanding}
\subsubsection{Reconstruction}
\label{sec:Reconstruction}
Scene reconstruction in robotics refers to the process of constructing a 3D representation of the environment using sensor data. 3DGS emerges as a promising scene representation to precisely model geometry and appearance information of the environment, enabling photo-realistic scene reconstruction. 3DGS-based reconstruction can be categorized into \textit{static reconstruction} and \textit{dynamic reconstruction}, depending on whether dynamic objects in the environment are modeled. We present a timeline of related works in Fig.~\ref{fig:recon_time}.

\begin{table*}[!t]
    \scriptsize 
    \caption{Categorization: 3DGS for Static Reconstruction.}
    \vspace{-0.05in}
    \label{tab:Static Reconstruction}
    \centering
    \resizebox{\linewidth}{!}{
    \begin{tabular}{l|ccc|cc|c|ccc}
    \toprule
    \multirow{2}{*}{Methods} & \multicolumn{3}{c|}{Input} & \multicolumn{2}{c|}{Appearance Modeling} & Geometry & \multicolumn{3}{c}{Scene Representation}\\
    \cline{2-10}
     & Monocular & Multi-camera & LiDAR & Illumination & Exposure & Normal & Single Gaussian \ \ \ \ & Hybrid Gaussian & Neural Field Gaussian\\
    \hline
    \rowcolor{mydark_blue}
    \multicolumn{10}{c}{Indoor Reconstruction}\\
    \hline
    360-GS~\cite{bai2024360} & \checkmark & & & & & & \checkmark & & \\
    \rowcolor{myblue}
    GaussianRoom~\cite{xiang2024gaussianroom} & \checkmark & & & & & \checkmark & \checkmark & & \\
    IM3DG~\cite{kim2024integrating} & \checkmark & & & & & & \checkmark & & \\
    \hline
    \rowcolor{mydark_blue}
    \multicolumn{10}{c}{Outdoor Reconstruction}\\
    \hline
    GaussianPro~\cite{cheng2024gaussianpro} & \checkmark & & & & & \checkmark & \checkmark & & \\
    \rowcolor{myblue}
    SWAG~\cite{dahmani2024swag} & \checkmark & & & \checkmark & & & \checkmark & & \\
    GauStudio~\cite{ye2024gaustudio} & \checkmark & & & \checkmark & & \checkmark & & \checkmark & \\
    \rowcolor{myblue}
    HGS-Mapping~\cite{wu2024hgs} & \checkmark & & \checkmark & & & & & \checkmark & \\
    HO-Gaussian~\cite{li2024ho} & & \checkmark & & \checkmark & & & & & \checkmark \\
    \rowcolor{myblue}
    TCLC-GS~\cite{zhao2024tclc} & \checkmark & & \checkmark & & & & & & \checkmark \\
    PGSR~\cite{chen2024pgsr} & \checkmark & & & & \checkmark & \checkmark & \checkmark & & \\
    \rowcolor{myblue}
    Wild-GS~\cite{xu2024wild} & \checkmark & & & \checkmark & \checkmark & & & & \checkmark\\
    WildGaussians~\cite{kulhanek2024wildgaussians} & \checkmark & & & \checkmark & & & & & \checkmark \\
    \rowcolor{myblue}
    DHGS~\cite{shi2024dhgs} & & \checkmark & \checkmark & & & & & \checkmark & \\
    \toprule
    \end{tabular}}
    \vspace{-0.1in}
\end{table*}

\noindent\textbf{Static Reconstruction.}\hspace{5pt} 
Static scene reconstruction focuses on environments with time-invariant geometry properties.
Considering that indoor and outdoor environments present distinct challenges in reconstruction tasks due to differences in scale, structural complexity, and lighting conditions, we split our discussion into indoor and outdoor scene reconstruction. 
Moreover, the effectiveness of static reconstruction methods heavily relies on the input sensor data, appearance and geometry modeling of the scenes, and the chosen scene representation. These factors play a crucial role in determining the accuracy, level of detail, and efficiency of the reconstruction process. Therefore, we categorize existing static reconstruction methods based on four criteria:  \textit{(i)} the type of input sensors, \textit{(ii)} the approach to appearance modeling, \textit{(iii)} the use of geometric normal constraint, \textit{(iv)} scene representation methods, as presented in Table \ref{tab:Static Reconstruction}. Particularly, appearance modeling includes illumination and exposure, as changing illumination conditions of the scenes and camera exposure problem lead to inaccurate appearance modeling of the environment. Scene representation in reconstruction can be classified into using a single 3D Gaussian representation (Single Gaussian), combining multiple Gaussian representations for joint scene modeling (Hybrid Gaussian), and integrating 3D Gaussian with neural network (Neural Field Gaussian).

Indoor scenes exhibit defined spatial layouts and rich textures. 
360-GS~\cite{bai2024360} introduces a layout-guided regularization on 3D Gaussians to reduce floaters caused by under-constrained regions. For modeling input panoramas, it projects 3D Gaussians onto the tangent plane of the unit sphere and then maps them to the spherical projections.
GaussianRoom~\cite{xiang2024gaussianroom} incorporates a learnable neural SDF field to guide the densification and pruning of Gaussian, achieving accurate surface reconstruction. Kim {\it et al.}~\cite{kim2024integrating} employ a hybrid representation for indoor scene reconstruction, using meshes to represent the room layout and 3D Gaussians for the reconstruction of other objects.

Outdoor scenes present additional challenges, such as varying illumination and scale variations. 
SWAG~\cite{dahmani2024swag} models the appearance of scenes using a learnable Multi-Layer Perceptron (MLP) network to address varying lighting conditions in outdoor environments. 
Wild-GS~\cite{xu2024wild} utilizes hierarchical appearance modeling by extracting high-level 2D appearance information and constructing the positional-aware local appearance embedding to handle the complex appearance variances across different views.
WildGaussians~\cite{kulhanek2024wildgaussians} enhances 3DGS with a per-Gaussian trainable embedding and uses a small MLP to integrate image and appearance embedding, which addresses varying illumination conditions in the wild.
PGSR~\cite{chen2024pgsr} proposes a camera exposure compensation model to address the large illumination variations in outdoor scenes.

Moreover, different parts of a scene often have distinct characteristics that require tailored modeling approaches. For example, sky regions, which are distant and lack geometric detail, are challenging to represent effectively with vanilla 3DGS in world coordinates. Similarly, modeling flat surfaces like roads with 3D Gaussians can lead to redundant representations.
To address these challenges, HGS-Mapping~\cite{wu2024hgs} uses a hybrid Gaussian representation: spherical Gaussians for sky modeling, 2D Gaussian planes for roads, and 3D Gaussians for roadside landscapes. 
Similarly, GauStudio~\cite{ye2024gaustudio} also uses spherical Gaussian maps to model the sky independently from the foreground. For road surfaces, DHGS~\cite{shi2024dhgs} employs 2DGS~\cite{huang20242d} with SDF-based surface constraints to enhance optimization. Additionally, HO-Gaussian~\cite{li2024ho} integrates a grid-based volume into the 3DGS pipeline to support geometric learning for the sky, distant, and low-texture areas. TCLC-GS~\cite{zhao2024tclc} combines colorized 3D mesh and hierarchical octree features to enhance the 3D Gaussian representation of urban scenes, while 
GaussianPro~\cite{cheng2024gaussianpro} applies planar loss to regularize Gaussian geometry in outdoor environments.



Since real-world scenarios are inherently complex and variable, accurately modeling appearance and geometry is essential for static reconstruction. For representing static scenes, hybrid Gaussians provide more precise modeling of the entire scene compared to single Gaussians, as they can adapt different modeling techniques to match the geometric structures of specific regions, whereas single Gaussians apply a uniform approach throughout. Additionally, neural field Gaussians excel in illumination modeling over both single and hybrid Gaussians by incorporating an MLP network, which enhances their capability to capture detailed lighting variations.

\begin{table*}[!t]
    \caption{Categorization: 3DGS for Dynamic Reconstruction.}
    \vspace{-0.07in}
    \label{tab:Dynamic Reconstruction}
    \centering
    \resizebox{\linewidth}{!}{
    \begin{tabular}{l|ccc|ccc|ccc}
    \toprule
    \multirow{2}{*}{Methods} & \multicolumn{3}{c|}{Input} & \multicolumn{3}{c|}{Dynamic-Static Separation} & \multicolumn{3}{c}{Dynamic Modeling}\\
    \cline{2-4}
    \cline{5-10}
     & Monocular & Multi-camera & LiDAR & 3D Prior & 2D Prior & No Prior & Time-varying & Deformation & 4D Gaussian \\
    \hline
    \rowcolor{myblue}
    PVG~\cite{chen2023periodic} & \checkmark & & \checkmark & & & \checkmark & \checkmark & & \\
    DrivingGaussian~\cite{zhou2024drivinggaussian} & & \checkmark & \checkmark & \checkmark & \checkmark & & \checkmark & & \\
    \rowcolor{myblue}
    Street Gaussians~\cite{yan2024street} & \checkmark & & \checkmark & \checkmark & \checkmark &  & \checkmark & & \\
    HUGS~\cite{zhou2024hugs} & \checkmark & & & \checkmark & \checkmark & & \checkmark & & \\
    \rowcolor{myblue}
    MoSca~\cite{lei2024mosca} & \checkmark & & & & \checkmark & & & \checkmark & \\
    $S^3$Gaussian~\cite{huang2024textit} & \checkmark & & \checkmark & & \checkmark & & & \checkmark & \\
    \rowcolor{myblue}
    4DGF~\cite{fischer2024dynamic} & \checkmark & & \checkmark & \checkmark & & & & \checkmark & \\
    VDG~\cite{li2024vdg} & & \checkmark & & & \checkmark & & \checkmark & & \\
    \rowcolor{myblue}
    EgoGaussian~\cite{zhang2024egogaussian} & \checkmark & & & & \checkmark & & & \checkmark & \\
    AutoSplat~\cite{khan2024autosplat} & \checkmark & & \checkmark & \checkmark & \checkmark & & & \checkmark & \\
    \rowcolor{myblue}
    Dynamic 3DG~\cite{luiten2023dynamic} & & \checkmark & & & & \checkmark & \checkmark & & \\
    Deformable 3DG~\cite{yang2024deformable} & \checkmark & & & & & \checkmark & & \checkmark & \\
    \rowcolor{myblue}
    4DGaussians~\cite{wu20244d} & \checkmark & & & & & \checkmark & & \checkmark & \\
    P4DGS~\cite{yang2023real} & \checkmark & & & & & \checkmark & & & \checkmark \\
    \rowcolor{myblue}
    Efficient 3DG~\cite{katsumata2023efficient} & \checkmark & & & & & \checkmark & \checkmark & & \\
    SC-GS~\cite{huang2024sc} & \checkmark & & & & & \checkmark & & \checkmark & \\
    \rowcolor{myblue}
    Gaussian-Flow~\cite{lin2024gaussian} & \checkmark & & & & & \checkmark & & \checkmark & \\
    GauFRe~\cite{liang2023gaufre} & \checkmark & & & & & \checkmark & & \checkmark & \\
    \rowcolor{myblue}
    SWAGS~\cite{shaw2023swags} & \checkmark & & & & & \checkmark & & \checkmark & \\
    STG~\cite{li2024spacetime} & \checkmark & & & & & \checkmark & \checkmark & & \\
    \rowcolor{myblue}
    4D-Rotor~\cite{duan20244d} & \checkmark & & & & & \checkmark & & & \checkmark \\
    3DGStream~\cite{sun20243dgstream} & \checkmark & & & & & \checkmark & & \checkmark & \\
    \rowcolor{myblue}
    ED3DGS~\cite{bae2024per} & \checkmark & & & & & \checkmark & & \checkmark & \\
    GaGS~\cite{lu20243d} & \checkmark & & & & & \checkmark & & \checkmark & \\
    \rowcolor{myblue}
    R3DG~\cite{zhang2024refined} & \checkmark & & & & & \checkmark & & \checkmark & \\
    SP-GS~\cite{wan2024superpoint} & \checkmark & & & & & \checkmark & & \checkmark & \\
    \rowcolor{myblue}
    AmbientGaussian~\cite{shih2024modeling} & \checkmark & & & & & \checkmark & & \checkmark & \\
    DGM~\cite{stearns2024dynamic} & \checkmark & & & & & \checkmark & \checkmark & & \\
    \rowcolor{myblue}
    GS-LK~\cite{xie2024gaussian} & \checkmark & & & & & \checkmark & & \checkmark & \\
    S4D~\cite{he2024s4d} & \checkmark & & & & & \checkmark & & \checkmark &\\
    \toprule
    \end{tabular}}
    \vspace{-0.1in}
\end{table*}
\begin{figure}[!t]
\centering
    
    \begin{subfigure}[b]{0.48\linewidth}
    \centering
    \includegraphics[width=\linewidth]{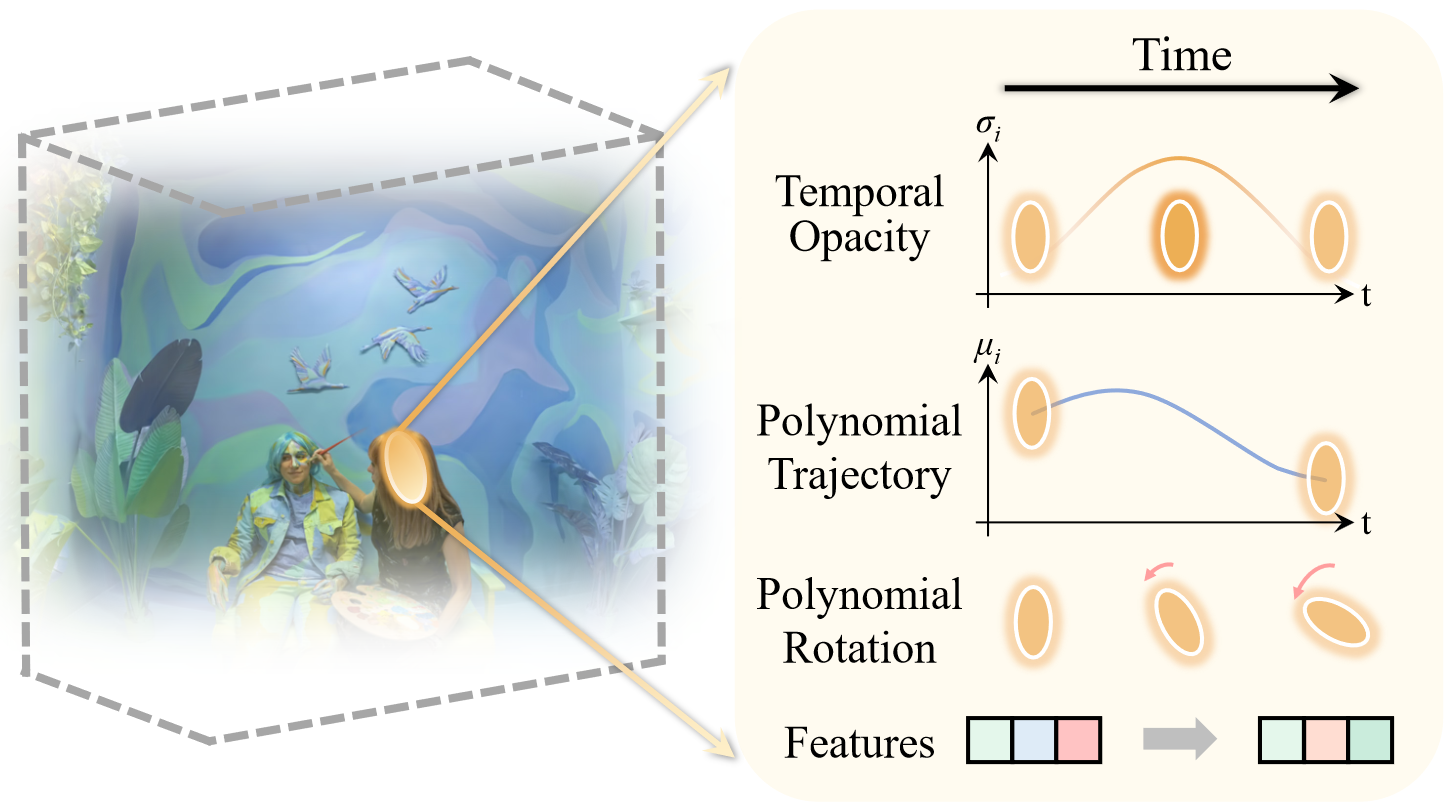}
    \caption{Time-varying-based dynamic reconstruction}
    \label{fig:dynamic_time_varying}
    \end{subfigure}
    \hfill
    \begin{subfigure}[b]{0.48\linewidth}
    \centering
    \includegraphics[width=\linewidth]{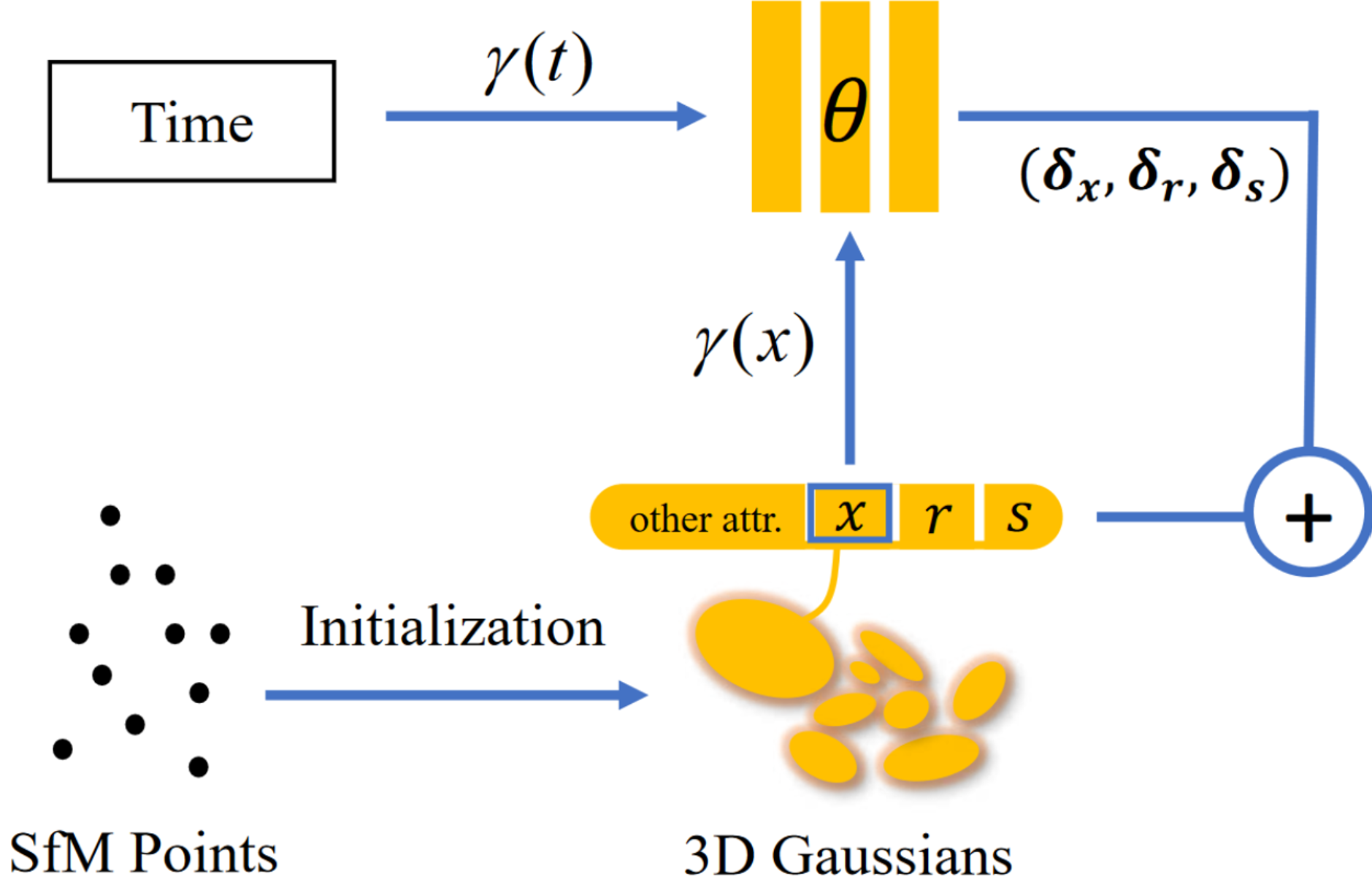}
    \caption{Deformation-based dynamic reconstruction}
    \label{fig:dynamic_deformation}
    \end{subfigure}
    \vspace{0.05in}

    \begin{subfigure}{\linewidth}
    \centering
    \includegraphics[width=\linewidth]{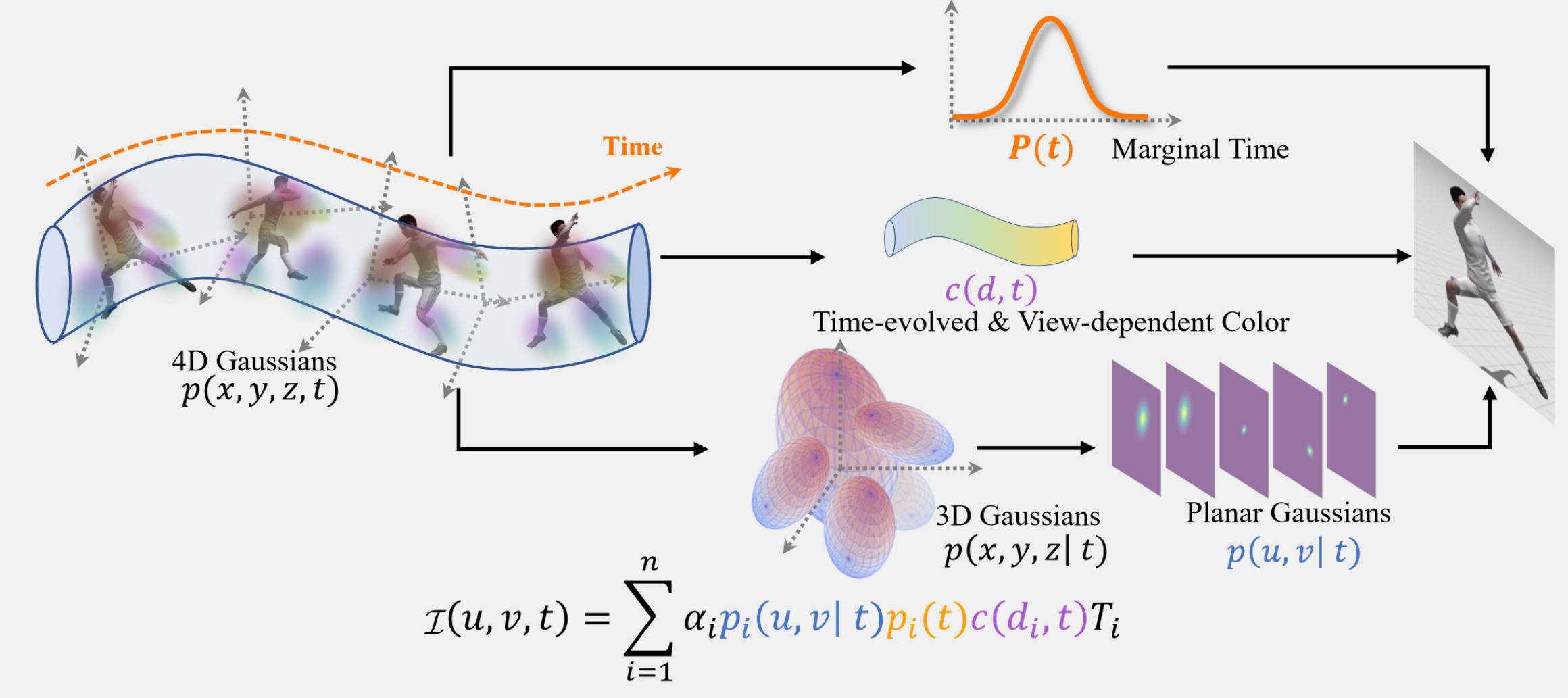}
    \caption{4D Gaussian-based dynamic reconstruction}
    \label{fig:dynamic_4dgs}
    \end{subfigure}
\vspace{-0.15in}
\caption{An illustration of 3DGS for dynamic reconstruction. Fig. \ref{fig:dynamic_time_varying}, Fig. \ref{fig:dynamic_deformation}, and Fig. \ref{fig:dynamic_4dgs} are originally shown in \cite{li2024spacetime}, \cite{yang2024deformable}, and \cite{yang2023real}, respectively.}
\vspace{-0.15in}
\label{fig:dynamic_reconstruction}
\end{figure}
\noindent\textbf{Dynamic Reconstruction.}\hspace{5pt} 
In real-world robotic applications, the presence of dynamic objects is inevitable. Vanilla 3DGS representation, primarily designed for static modeling, struggles with modeling the motions of dynamic objects as they inevitably interfere with the optimization of Gaussian parameters. Therefore, learning 3DGS-based models in dynamic scenes is crucial. For dynamic reconstruction, the key issues are how to differentiate between dynamic and static components, and how to model the dynamic objects.

Therefore, we categorize existing 3DGS-based dynamic reconstruction methods based on three criteria: \textit{(i)} the type of input sensors, \textit{(ii)} methods to separate static and dynamic objects in the scenes, \textit{(iii)} the dynamic modeling methods, as presented in Table \ref{tab:Dynamic Reconstruction}. Specifically, the separation of static and dynamic components in the scenes can be achieved by applying motion or semantic masks to detect the locations of dynamic objects in images (referred to as 2D prior), using 3D bounding boxes to identify dynamic objects (3D prior), or by simultaneously modeling both static and dynamic elements (no prior). Dynamic modeling methods can be classified into \textit{time-varying}, \textit{deformation-based}, and \textit{4D Gaussian} modeling, as shown in Fig.~\ref{fig:dynamic_reconstruction}.

In terms of time-varying modeling methods, each 3D Gaussian position is expressed as a function of time to model the temporal change of the position. Dynamic 3DG~\cite{luiten2023dynamic} regularizes Gaussians’ motion and rotation with local-rigidity constraints.
PVG~\cite{chen2023periodic} and VDG~\cite{li2024vdg} introduce periodic vibration-based temporal dynamics by modifying the mean and opacity of vanilla 3DGS to be time-dependent functions centered around the life peak. DrivingGaussian~\cite{zhou2024drivinggaussian} and Street Gaussians~\cite{yan2024street} leverage a composite dynamic Gaussian graph to represent multiple moving objects across time. HUGS~\cite{zhou2024hugs} extends 3DGS to model camera exposure on dynamic scenes and utilizes optical flow prediction for dynamic separation. Efficient 3DG~\cite{katsumata2023efficient} models the temporal changes in position dynamics using Fourier approximation. STG~\cite{li2024spacetime} stores features instead of SH coefficients in each Gaussian to accurately encode view- and time-dependent radiance. 

Deformation-based methods model motions as deformations relative to the canonical space of previous observations that represent a static field. 
$S^3$Gaussian~\cite{huang2024textit} and 4DGaussians~\cite{wu20244d} employ a multi-resolution spatial-temporal field network to represent dynamic 3D scenes following Hexplane~\cite{cao2023hexplane} and an lightweight MLP to decode deformation. 4DGF~\cite{fischer2024dynamic} utilizes neural fields to represent sequence- and object-specific appearance and geometry variations. AutoSplat~\cite{khan2024autosplat} estimates residual SH for each foreground Gaussian to model the dynamic appearance of foreground objects. 
SC-GS~\cite{huang2024sc} introduces sparse control points together with an MLP for modeling scene motion.
Gaussian-Flow~\cite{lin2024gaussian} models time-dependent residuals of each Gaussian attribute by a polynomial fitting in the time domain, and a Fourier series fitting in the frequency domain. 
SWAGS~\cite{shaw2023swags} partitions the sequence into windows and trains a separate dynamic 3DGS model for each window, allowing the canonical representation to change. 
3DGStream~\cite{sun20243dgstream} employs neural transformation cache (NTC) to model the translations and rotations of 3DGS. 
ED3DGS~\cite{bae2024per} introduces coarse and fine temporal embeddings to represent the slow and fast state of the dynamic scene.
GaGS~\cite{lu20243d} extracts 3D geometry features and integrates them in learning the 3D deformation.
GS-LK~\cite{xie2024gaussian} introduces an analytical regularization of the warp canonical field in dynamic 3DGS by deriving a Lucas-Kanade style velocity field. S4D~\cite{he2024s4d} employs partially learnable control points for local 6-DoF motion representation.

4D Gaussian methods consider the spacetime as an entirety and incorporate a time dimension into 3D Gaussian primitives, forming 4D Gaussian primitives for dynamic modeling.
P4DGS~\cite{yang2023real} employs a 4D Gaussian parameterized by anisotropic ellipses that can rotate arbitrarily in space and time, as well as view-dependent and time-evolved appearance represented by 4D SH coefficients. Instead of using isoclinic quaternions for 4D rotation representation as in P4DGS~\cite{yang2023real}, 4D-Rotor~\cite{duan20244d} introduces 4D rotors to characterize the 4D rotations motivated by \cite{bosch2020n}, achieving decoupled spatial and temporal rotations.

Current 3DGS dynamic modeling methods have demonstrated their capability to reconstruct dynamic scenes. In small-scale scenes, these methods can reconstruct both dynamic and static components in a unified manner. However, when dealing with larger-scale autonomous driving scenarios, the computational burden of unified  reconstruction becomes increasingly high. To address this issue, additional prior information is required to distinguish between dynamic and static regions, which are then reconstructed separately. The limitation of these methods lies in their reliance on additional prior information, such as 3D bounding boxes, which are not easily accessible. Consequently, the future development of dynamic reconstruction, besides enhancing the precision of dynamic modeling, is to leverage the geometry modeling of 3DGS for larger-scale dynamic reconstruction with minimal prior knowledge.

\begin{figure*}[!t]
\centering 
\includegraphics[width=\linewidth]{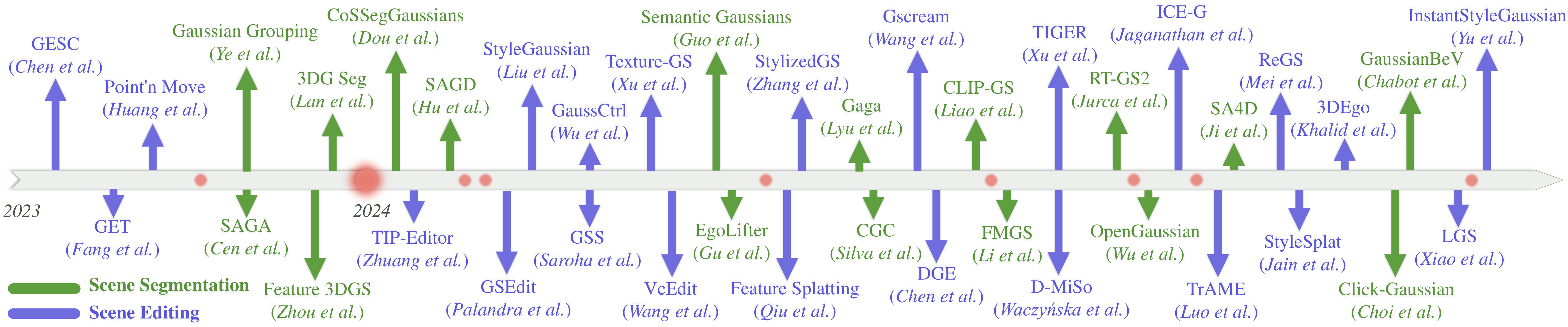}
\vspace{-0.2in}
\caption{Chronological: 3DGS for Scene Segmentation and Editing.}
\vspace{-0.05in}
\label{fig:seg_time}
\end{figure*}
\begin{table}[!t]
\caption{Categorization: 3DGS for Scene Segmentation.}
\vspace{-0.05in}
\label{tab:Semantic Segmentation}
\centering
\resizebox{\linewidth}{!}{
\begin{tabular}{l|ccc|ccc}
\toprule
\multirow{3}{*}{Methods} & \multicolumn{3}{c|}{Semantic representation} & \multicolumn{3}{c}{Multi-view Semantic Consistency}\\
\cline{2-7}
  &Semantic & Feature & \multirow{2}{*}{MLP} & \multirow{2}{*}{Prior} & Contrastive & 3D Spatial \\
  & Labels & Embedding & & & Learning & Association \\
\hline
 \rowcolor{myblue}
Gaussian Grouping~\cite{ye2023gaussian} & & \checkmark & & \checkmark & & \\
SAGA~\cite{cen2023saga} & & \checkmark & & & \checkmark & \\
\rowcolor{myblue}
Feature 3DGS~\cite{zhou2024feature} & & \checkmark & & & & \checkmark \\
3DG Seg~\cite{lan20232d} & \checkmark & & & \checkmark & & \\
\rowcolor{myblue}
CoSSegGaussians~\cite{dou2024cosseggaussians} & & & \checkmark & \checkmark & & \\
SAGD~\cite{hu2024semantic} & \checkmark & & & \checkmark & & \\
\rowcolor{myblue}
Semantic Gaussians~\cite{guo2024semantic} & \checkmark & & & & & \checkmark \\
EgoLifter~\cite{gu2024egolifter} & & \checkmark & & & \checkmark & \\
\rowcolor{myblue}
Gaga~\cite{lyu2024gaga} & & \checkmark & & & & \checkmark \\
CGC~\cite{silva2024contrastive} & & \checkmark & & & \checkmark & \\
\rowcolor{myblue}
CLIP-GS~\cite{liao2024clip} & & \checkmark & & \checkmark & & \\
FMGS~\cite{zuo2024fmgs} & & & \checkmark & & & \checkmark \\
\rowcolor{myblue}
RT-GS2~\cite{jurca2024rt} & & \checkmark & & & \checkmark & \\
OpenGaussian~\cite{wu2024opengaussian} & & \checkmark & & & \checkmark & \\
\rowcolor{myblue}
Click-Gaussian~\cite{choi2024click} & & \checkmark & & & \checkmark & \\
SA4D~\cite{ji2024segment} & & \checkmark & & \checkmark  & &\\
\rowcolor{myblue}
GaussianBeV~\cite{chabot2024gaussianbev} & & \checkmark & & \checkmark  & &\\
\toprule
  \end{tabular}}
\end{table}
\subsubsection{Segmentation \& Editing}
\label{sec:Segmentation}
The timeline of the scene segmentation and editing is illustrated in Fig.~\ref{fig:seg_time}.

\noindent\textbf{Scene Segmentation.}\hspace{5pt} 
Scene segmentation divides an observed scene into distinct components, each representing a unique semantic category. Unlike 2D segmentation, 3D segmentation better addresses the operational and navigational needs of robots in real-world environments. 3DGS provides an advanced scene representation framework that enables 3D semantic segmentation based on 2D images.
We categorize current 3DGS segmentation methods by two main criteria: \textit{(i)} the modeling approach for 3DGS semantic representation and \textit{(ii)} methods for maintaining consistent semantic labels across multiple 2D image inputs, as outlined in Table~\ref{tab:Semantic Segmentation}.
Specifically, semantic representation approaches fall into three categories: semantic labels, feature embeddings, and MLP-based models. The first two incorporate semantic labels and feature embeddings directly into 3D Gaussian primitives as additional attributes, while the third relies on an MLP network for semantic modeling. 

Most 3DGS segmentation methods currently depend on the SAM model~\cite{kirillov2023segment} to supply 2D semantic labels for open-vocabulary segmentation. However, since the SAM model ensures semantic consistency only within individual images, corresponding semantic regions across different images may not align consistently, leading to ambiguities in 3D semantic modeling. To address this, existing 3DGS methods use priors, contrastive learning, or 3D spatial association techniques to achieve multi-view semantic consistency in 3D semantic segmentation.
Prior-based methods employ pretrained tracker~\cite{cheng2023tracking} or input multi-view consistent 2D masks to propagate and associate masks across views.
Gaussian Grouping~\cite{ye2023gaussian} utilizes a 2D identity loss and a 3D regularization loss for Gaussian optimization, leveraging the coherent segmentation across views.
SAGD~\cite{hu2024semantic} incorporates a Gaussian decomposition module to address boundary roughness in 3D segmentation. 
CoSSegGaussians~\cite{dou2024cosseggaussians} designs a multi-scale spatial and semantic Gaussian features fusion module to achieve compact segmentation.
CLIP-GS~\cite{liao2024clip} introduces coherent semantic regularization by incorporating the semantics of adjacent views to eliminate semantic ambiguity within the same object. 
SA4D~\cite{ji2024segment} introduces a temporal identity feature field to learn Gaussians’ identity information across time for 4D segmentation.
GaussianBeV~\cite{chabot2024gaussianbev} transforms image features into bird's-eye view (BEV) Gaussian representation for BEV segmentation in an end-to-end manner.

Contrastive learning works align features of corresponding masks while separating features of different masks in the embedding space to achieve cross-view semantic consistency. For semantic gaussian optimization, SAGA~\cite{cen2023saga} introduces the correspondence loss based on the principle that pixels with a higher intersection over union (IoU) should have more similar features.
OpenGaussian~\cite{wu2024opengaussian} proposes an intra-mask smoothing loss and inter-mask contrastive loss to promote feature diversity among different instances. CGC~\cite{silva2024contrastive} presents a spatial-similarity regularization loss to enforce the spatial continuity of feature vectors, addressing misclassification in regions where the scene is not well observed. 
Moreover, RT-GS2~\cite{jurca2024rt} fuses view-dependent features extracted from images and view-independent 3D Gaussian features obtained from contrastive learning to enhance semantic consistency across different views.
Click-Gaussian~\cite{choi2024click} incorporates coarse-to-fine level features into Gaussian and clusters global features from noisy 2D segments across views to enhance global semantic consistency.

3D spatial association methods employ 3D semantic feature extraction network or label voting in spatial coordinates to achieve semantic consistency.
Semantic Gaussians~\cite{guo2024semantic} constructs loss between 3D features obtained from 3D semantic network and mapped 3D Gaussian features from 2D features projection to optimize 3DGS semantic representation.
Gaga~\cite{lyu2024gaga} employs 3D-aware mask association process, where a 3D-aware memory bank is used to assign a consistent group ID based on the overlap ratio of shared 3D Gaussians between each mask and existing groups in the memory bank.
FMGS~\cite{zuo2024fmgs} integrates semantic and language representation for scene understanding by distilling vision-language features from foundation models CLIP~\cite{radford2021learning} and DINO~\cite{caron2021emerging} into 3DGS. Additionally, this work enhances spatial precision through a pixel alignment loss leveraging DINO features.

In summary, 3DGS enables faster and more accurate 3D semantic modeling for semantic segmentation compared to other scene representations. This improvement stems from the Gaussian radiance field representation in 3DGS, which supports detailed scene modeling, and its efficient rendering capability, which accelerates optimization. Additionally, the explicit structure of Gaussians and 3DGS-based semantic segmentation results make it straightforward to edit and manipulate semantic objects within the scene representation. This functionality enhances 3DGS’s applicability in various downstream robotic tasks, such as manipulation and autonomous navigation, where understanding and interacting with semantic entities in the environment are essential.

\begin{figure}[!t]
\centering
    \begin{subfigure}{\linewidth}
    \centering
    \includegraphics[width=\linewidth]{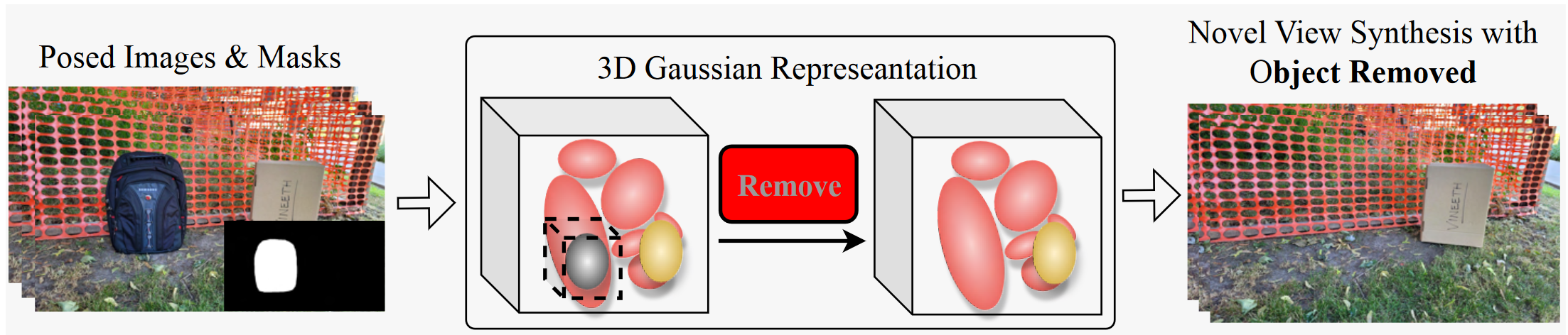}
    \caption{Object editing}
    \vspace{0.05in}
    \label{fig:edit_object}
    \end{subfigure}
    
    \begin{subfigure}{\linewidth}
    \centering
    \includegraphics[width=\linewidth]{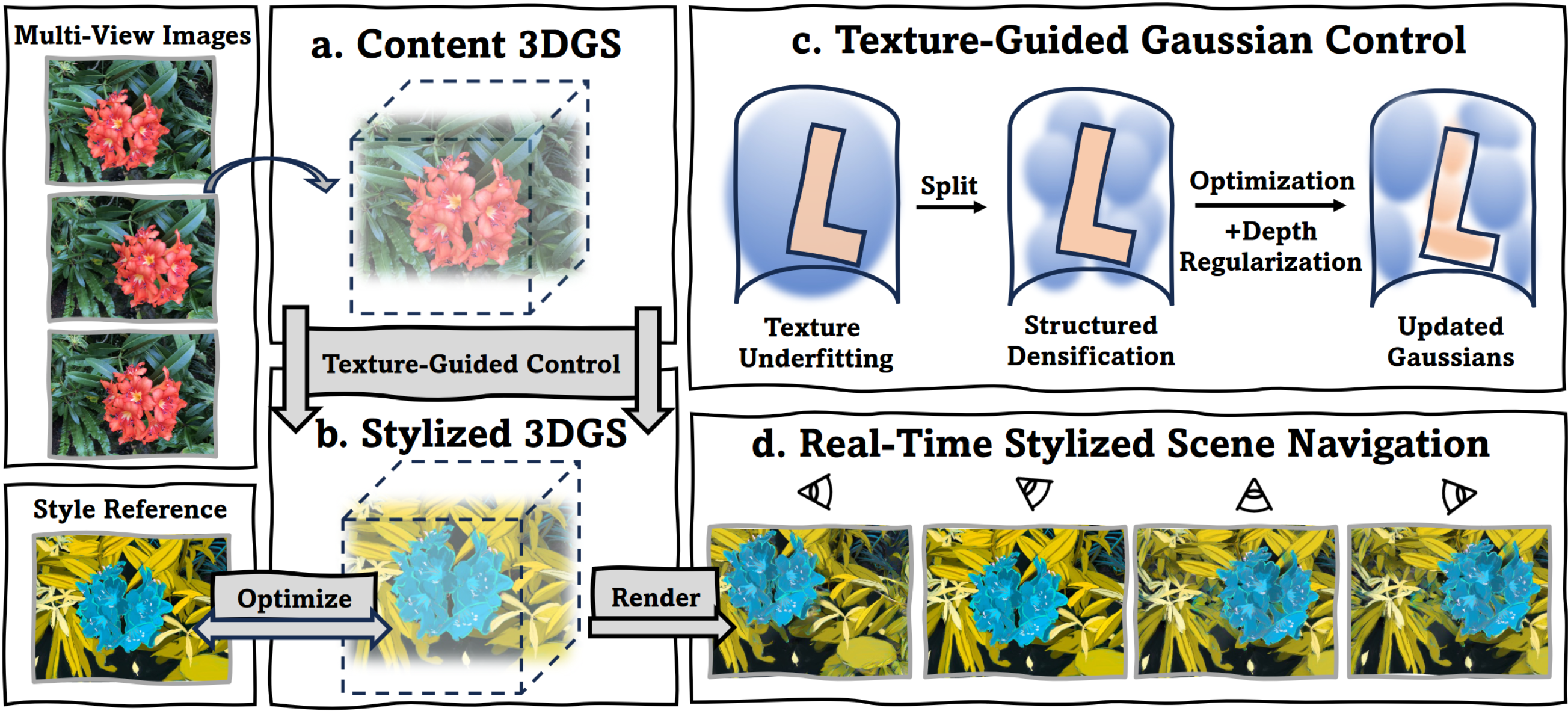}
    \caption{Scene style editing}
    \label{fig:edit_style}
    \end{subfigure}

\vspace{-0.05in}
\caption{An illustration of 3DGS for scene editing. Fig. \ref{fig:edit_object} and Fig. \ref{fig:edit_style} are originally shown in \cite{wang2024gscream} and \cite{mei2024reference}, respectively.}
\vspace{-0.1in}
\label{fig:edit}
\end{figure}

\noindent\textbf{Scene Editing.}\hspace{5pt} 
Scene editing involves modifying scene elements based on user prompts to achieve desired effects. Edited scenes can serve as valuable training resources for robots, offering a practical alternative when real-world data collection is challenging or time-consuming. The 3DGS representation simplifies the editing process by utilizing the explicit structure of Gaussians, allowing for direct relocation of Gaussian elements to facilitate scene modifications. We categorize related works into \textit{object editing} and \textit{scene style editing}, as shown in Fig.~\ref{fig:edit}.

Object editing methods includes \textit{object insertion and removal}~\cite{wang2024gscream, huang2023point, luo20243d, qiu2024feature} as well as \textit{appearance and texture editing}~\cite{xu2024texture}. The challenges of object insertion and removal editing are the preservation of geometric consistency and the maintenance of texture coherence, due to the discrete properties of Gaussian primitives.
GScream~\cite{wang2024gscream} employs cross-attention feature regularization to propagate the accurate texture in the surrounding region into the in-painted region, ensuring texture coherence of the Gaussian representation. 
Feature Splatting~\cite{qiu2024feature} incorporates visual-language embedding into 3D Gaussian for text-prompt editing. For appearance and texture editing, Texture-GS~\cite{xu2024texture} establishes a connection between geometry (3D Gaussians) and appearance (2D texture map) to facilitate appearance editing.

Text-driven object editing methods~\cite{zhuang2024tip, palandra2024gsedit, wu2024gaussctrl, wang2024view, wang2024gaussianeditor, chen2024gaussianeditor, chen2024dge, xu2024tiger, luo2024trame, khalid20243dego, xiao2024localized} all leverage pretrained diffusion model~\cite{rombach2022high}, which enables efficient use of text prompts for precise editing and incorporates rich prior knowledge to enhance the quality and coherence of the edited objects.
GaussCtrl~\cite{wu2024gaussctrl} employs naturally consistent depth maps obtained from ControlNet~\cite{zhang2023adding} to guide geometric consistency. Moreover, this work introduces self and cross-view attention of latent codes from different views for appearance consistency.
VcEdit~\cite{wang2024view} and TrAME~\cite{luo2024trame} employ cross-attention across maps obtained from stages of diffusion.

Scene style editing methods~\cite{liu2024stylegaussian, zhang2024stylizedgs, saroha2024gaussian, jaganathan2024ice, jain2024stylesplat, mei2024reference, yu2024instantstylegaussian} generate diverse stylistic scene data based on style prompts, which can shorten data collection periods and improve the robustness of the systems trained with this data for robotic applications.

\begin{figure*}[!t]
\centering 
\includegraphics[width=\linewidth]{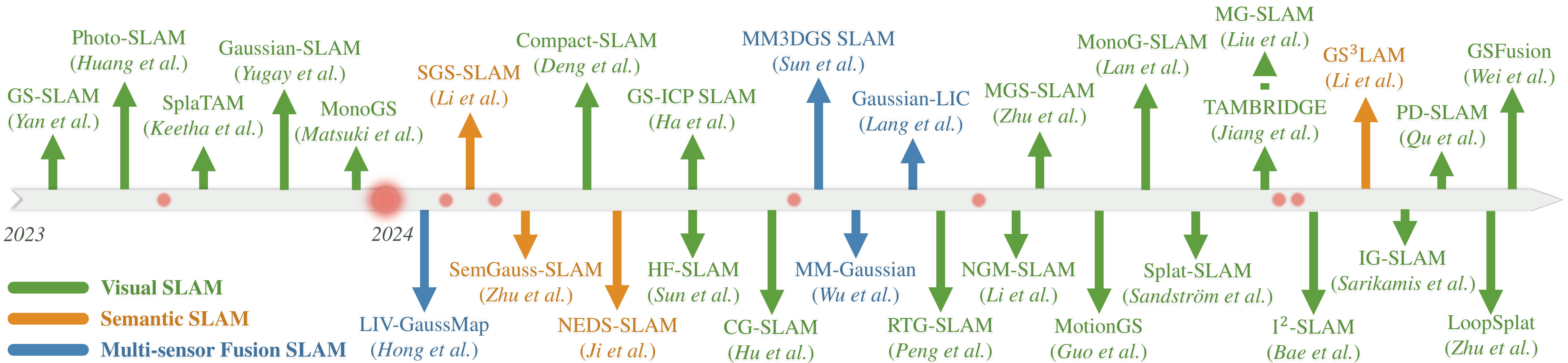}
\caption{Chronological: 3DGS for SLAM.}
\vspace{-0.1in}
\label{fig:slam_time}
\end{figure*}
\subsubsection{SLAM}
\label{sec:SLAM}
\begin{figure}[!t]
\centering 
\includegraphics[width=\linewidth]{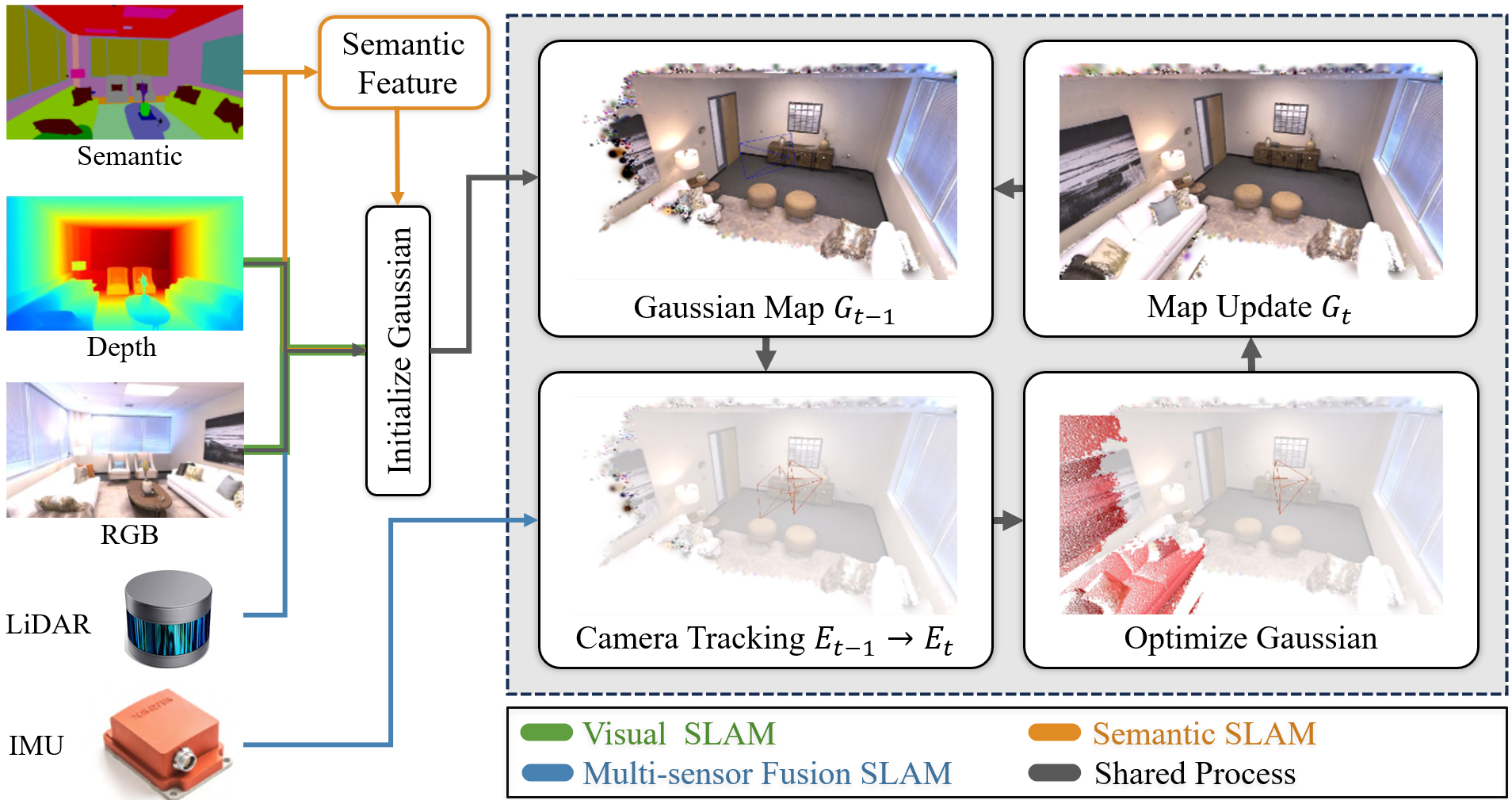}
\vspace{-0.1in}
\caption{An illustration of 3DGS for SLAM. Part of images are taken from SplaTAM~\cite{keetha2024splatam}.}
\vspace{-0.1in}
\label{fig:slam}
\end{figure}

We categorize 3DGS-based SLAM into \textit{visual SLAM}, \textit{multi-sensor fusion SLAM}, \textit{semantic SLAM} based on the sensor used and the level of environmental understanding. 3DGS-based SLAM process is illustrated in Fig.~\ref{fig:slam}. We also present timeline of the related works in Fig.~\ref{fig:slam_time}. Compared with NeRF-based SLAM, 3DGS-based SLAM methods leverage the explicit representation of 3D Gaussian for unbounded and photo-realistic mapping. 

\noindent\textbf{Visual SLAM.}\hspace{5pt} 
Visual SLAM refers to the process of simultaneously performing dense color mapping and localization of a camera. Considering that accurate depth information is crucial for providing geometric supervision in constructing dense Gaussian maps, we categorize our discussion of visual SLAM into RGB-D SLAM and RGB SLAM, based on whether the input camera data contains depth information.

RGB-D SLAM methods~\cite{keetha2024splatam, yugay2023gaussian, li2024ngm, liu2024structure, sun2024high, hu2024cg, yan2024gs, ha2024rgbd, jiang2024tambridge, deng2024compact, peng2024rtg, qu2024visual, bae20242, zhu2024loopsplat, wei2024gsfusion} have accurate depth input, which can provide good geometric constraints for dense mapping and pose estimation. For SLAM process, it is crucial to determine when to expand the map and how to achieve precise tracking results. SplaTAM~\cite{keetha2024splatam} employs silhouette rendering of the existing Gaussian map to identify which portions of the scene are new content, thereby guiding the expansion of the map and the optimization of camera poses. Gaussian-SLAM~\cite{yugay2023gaussian} and NGM-SLAM~\cite{li2024ngm} focus on building Gaussian submaps progressively to achieve local map optimization. MG-SLAM~\cite{liu2024structure} incorporates structure prior of the scene to generate constraints that address gaps and imperfections of the reconstructed map. 
In addition, CG-SLAM~\cite{hu2024cg} proposes a novel depth uncertainty model to build up a consistent and stable 3D Gaussian map. I$^2$-SLAM~\cite{bae20242} integrates the image formation process into SLAM to overcome motion blur and varying appearances. LoopSplat~\cite{zhu2024loopsplat} detects loop closure online and computes relative loop edge constraints between submaps via 3DGS registration for global map consistency.

    



For accurate tracking, GS-SLAM~\cite{yan2024gs} employs a coarse-to-fine approach to avoid drifted camera tracking caused by artifacts in images. GS-ICP SLAM~\cite{ha2024rgbd} leverages 3D explicit representation of 3DGS for tracking, achieved by utilizing G-ICP~\cite{segal2009generalized} for Gaussian map matching to directly regress estimated camera poses. TAMBRIDGE~\cite{jiang2024tambridge} jointly optimizes sparse re-projection and dense rendering errors, resulting in reduced cumulative errors in tracking.
While these advancements improve SLAM accuracy, detailed 3D Gaussian maps generated in the SLAM process require significant memory resources. To address this, Compact-SLAM~\cite{deng2024compact} proposes a sliding window-based online masking method to remove redundant Gaussian ellipsoids for compact representation. RTG-SLAM~\cite{peng2024rtg} uses a single opaque Gaussian instead of multiple overlapping Gaussians to fit a local region of the surface for reducing memory consumption in mapping. 

RGB SLAM methods~\cite{matsuki2024gaussian, guo2024motiongs, huang2024photo, zhu2024mgs, lan2024monocular, sandstrom2024splat, sarikamis2024ig} lack accurate depth input, so it requires additional information such as multi-view constraints or depth estimation to recover the 3D geometry of the scene. MonoGS~\cite{matsuki2024gaussian} and MotionGS~\cite{guo2024motiongs} employ a regularization term and multi-view optimization to constrain scene geometry information in scenes where depth is unknown. Photo-SLAM~\cite{huang2024photo} minimizes the reprojection error between matched 2D geometric keypoints in the frame and corresponding 3D points to achieve geometric consistency. Moreover, MGS-SLAM~\cite{zhu2024mgs} utilizes depth estimation and DPVO~\cite{teed2024deep} to recover 3D geometry of the scene. MonoG-SLAM~\cite{lan2024monocular} employs patch graph based on DPVO~\cite{teed2024deep} and CLIP~\cite{radford2021learning}-based loop closure optimization to guide the estimation of the scene geometry. Splat-SLAM~\cite{sandstrom2024splat} introduces dense optical flow estimation and DSPO (disparity, scale and pose optimization) for accurate depth recovery. 

The accuracy of existing 3DGS-based visual SLAM is highly dependent on precise depth information. RGB SLAM methods, due to the absence of accurate depth information, often exhibit errors in 3DGS geometric reconstruction. Although depth estimation techniques can provide depth information for RGB SLAM, their limited accuracy leads to decreased SLAM performance. Moreover, current 3DGS-based visual SLAM systems are tested and evaluated mostly in indoor environments, as depth measurements are unreliable in outdoor scenes. Therefore, the unsolved key problem of visual SLAM is to enhance geometric reconstruction accuracy despite inaccurate depth information, thereby achieving high-precision SLAM in various environments.

\noindent\textbf{Multi-sensor Fusion SLAM.}\hspace{5pt} 
Multi-sensor Fusion SLAM integrates data from different sensors to achieve accurate mapping and robust tracking in SLAM system. We classify multi-sensor fusion SLAM methods into \textit{LiDAR-based} and \textit{image depth estimation-based}, depending on the primary sensor modality used for 3DGS geometry reconstruction. 

LiDAR-based methods use accurate point cloud obtained from LiDAR as the initial input for 3D Gaussian geometry representation.
LIV-GaussMap~\cite{hong2024liv} utilizes point cloud obtained from an IESKF-based LiDAR-inertial system to provide an initial Gaussian structure for the scene.
Gaussian-LIC~\cite{lang2024gaussian} performs LiDAR-Inertial-Camera odometry for tracking. LiDAR points are projected onto the corresponding image, where they are colored by querying the pixel values, and then used to initialize 3D Gaussian.
MM-Gaussian~\cite{wu2024mm} performs tracking using point cloud registration algorithm~\cite{vizzo2023kiss} and utilizes multi-frame camera constraints for relocalization when the tracking module fails.
Image depth estimation-based methods employ monocular dense depth estimation results as geometric supervision for constructing Gaussian map. 
MM3DGS SLAM~\cite{sun2024mm3dgs} utilizes IMU (Inertial Measurement Unit) pre-integration for initial pose estimation and constructs loss between rendering and observations for pose optimization. As depth estimation outputs a relative depth, Pearson correlation coefficient is used to compute the depth loss between estimated and rendered depth maps with actual scale information.

\noindent\textbf{Semantic SLAM.}\hspace{5pt} 
Semantic SLAM incorporates semantic understanding of the environment into map construction and estimates camera pose simultaneously. Compared with visual SLAM and multi-sensor fusion SLAM, semantic SLAM enables dense semantic mapping of the scene, which is essential for downstream tasks such as navigation and manipulation. As the original 3DGS representation lacks semantic information, two methods have been developed in 3DGS-based semantic SLAM to incorporate semantics: color-based and feature-based. For color-based method, SGS-SLAM~\cite{li2024sgs} utilizes semantic color associated with the Gaussian for semantic representation. However, this color-based semantic modeling approach overlooks the higher-level information inherent in semantics.
For feature-based semantic integration, SemGauss-SLAM~\cite{zhu2024semgauss} incorporates semantic embedding into 3D Gaussian for semantic representation and performs semantic-informed bundle adjustment using multi-view constraints to achieve high-accuracy semantic SLAM. 
NEDS-SLAM~\cite{ji2024neds} proposes a fusion module that combines semantic features with appearance features to address the spatial inconsistency of semantic features, inspired by~\cite{zhu2024sni}.  
GS$^3$LAM~\cite{li2024gs} introduces depth-adaptive scale regularization to reduce the blurring of geometric surfaces induced by irregular Gaussian scales within semantic gaussian.
LEGS~\cite{yu2024legs} introduces a language-embedded Gaussian splat representation for semantic mapping.

\begin{figure}[!t]
\centering 
\includegraphics[width=\linewidth]{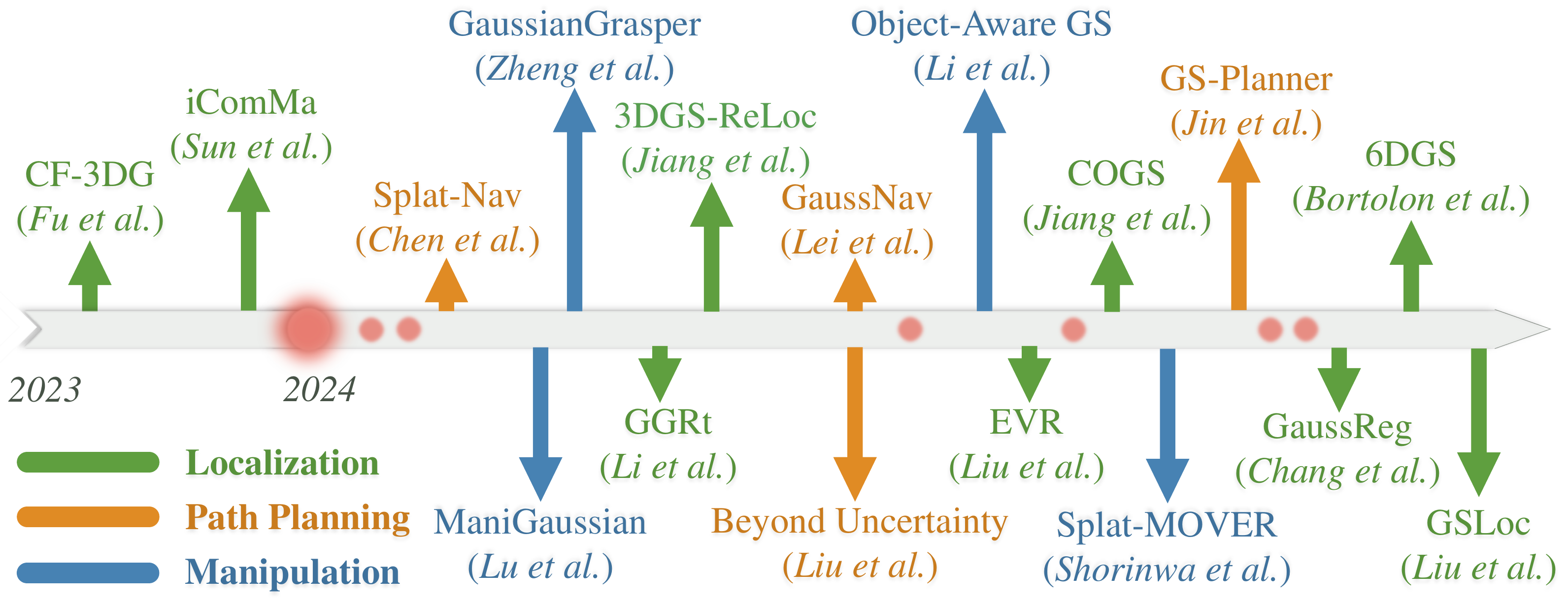}
\vspace{-0.2in}
\caption{Chronological: 3DGS for Scene Interaction.}
\vspace{-0.1in}
\label{fig:interaction_time}
\end{figure}
\subsection{Scene Interaction}
\label{sec:Scene Interaction}
Navigation and manipulation are fundamental aspects of robotic interaction with the environment. The timeline for related works is shown in Fig.~\ref{fig:interaction_time}.

\subsubsection{Manipulation}
\label{sec:Manipulation}
\begin{figure}[!t]
\centering
    \begin{subfigure}{\linewidth}
    \centering
    \includegraphics[width=\linewidth]{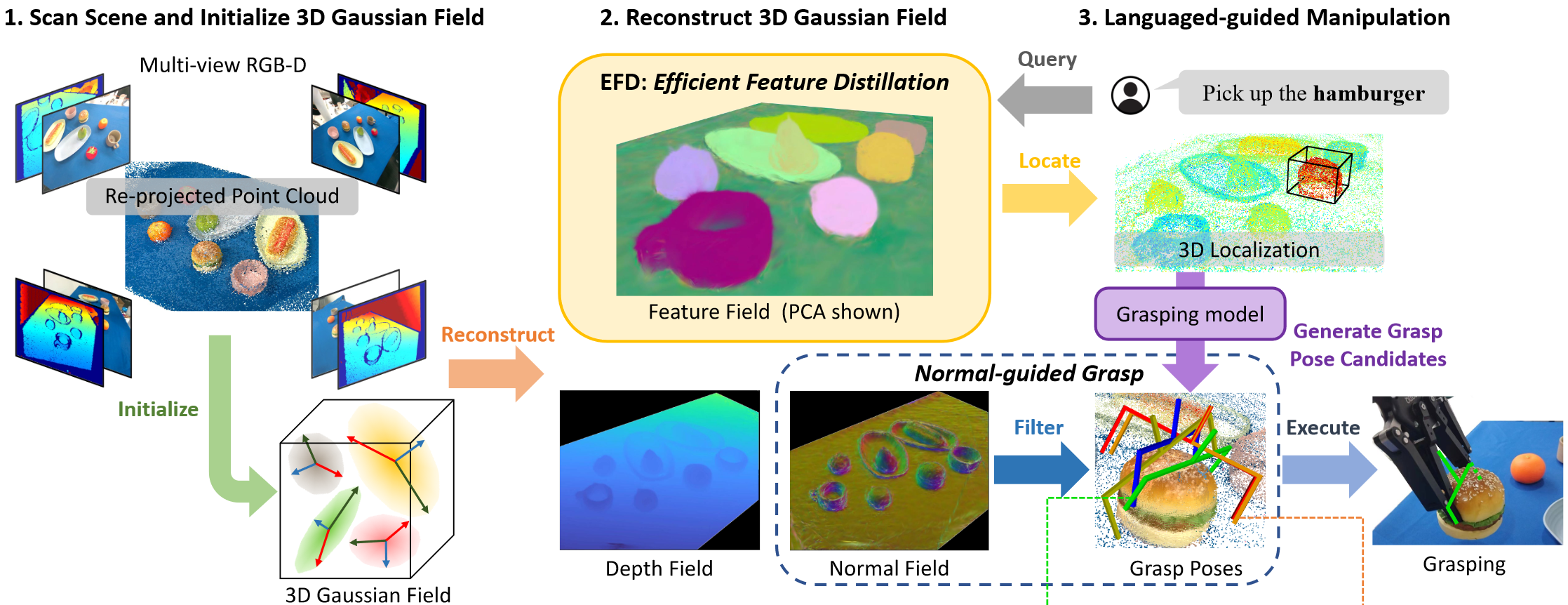}
    \caption{Single-stage manipulation}
    \vspace{0.05in}
    \label{fig:manipulation_single}
    \end{subfigure}
    
    \begin{subfigure}{\linewidth}
    \centering
    \includegraphics[width=\linewidth]{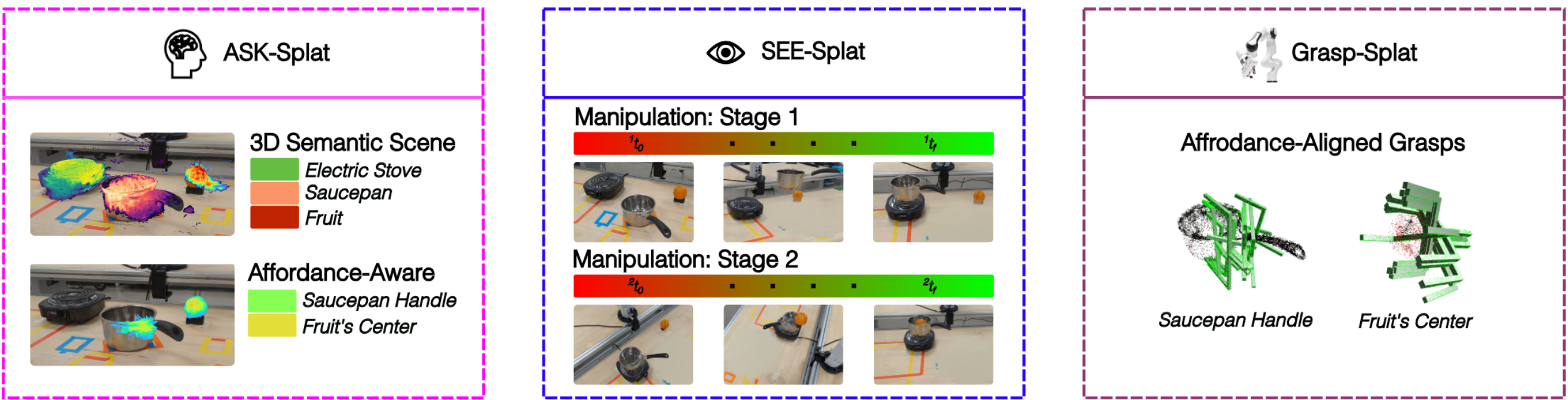}
    \caption{Multi-stage manipulation}
    \label{fig:manipulation_multi}
    \end{subfigure}
\vspace{-0.2in}
\caption{An illustration of 3DGS for manipulation. Fig. \ref{fig:manipulation_single} and Fig. \ref{fig:manipulation_multi} are originally shown in \cite{zheng2024gaussiangrasper} and \cite{shorinwa2024Splat}, respectively.}
\vspace{-0.1in}
\label{fig:manipulation}
\end{figure}

Manipulation refers to using robotic arms or grippers to perform various tasks, substituting for human hands. Unlike NeRF-based manipulation approaches, 3DGS employs an explicit radiance field representation which constructs the scene and provides direct access to the positional information of objects within it. 
Manipulation tasks can be categorized into \textit{single-stage} and \textit{multi-stage} based on whether the task requires consideration of dynamic environmental changes. These changes are caused by object movement resulting from the manipulation itself, as illustrated in Fig.~\ref{fig:manipulation}. 

In single-stage manipulation, the grasping task is completed through a single, continuous motion, so the environment is considered static during the process. GaussianGrasper~\cite{zheng2024gaussiangrasper} reconstructs a 3D Gaussian feature field via efficient feature distillation to support language-guided manipulation tasks and uses rendered normals to filter out unfeasible grasp poses. 

For multi-stage manipulation, the task is accomplished through a sequence of actions where the object's movement during each stage results in dynamic changes to the environment.
ManiGaussian~\cite{lu2024manigaussian} proposes dynamic 3DGS framework to model the scene-level spatiotemporal dynamics and builds a Gaussian world model to parameterize distributions in dynamic 3DGS model for multi-task robotic manipulation. 
Object-Aware GS~\cite{li2024object} models the Gaussian representation as time-variant and employs object-centric dynamic updates for dynamic modeling of multi-stage manipulation.
Instead of modeling dynamic changes, Splat-MOVER~\cite{shorinwa2024Splat} employs a scene-editing module using 3D semantic masking and infilling to visualize the motions of the objects that result from the robot's interactions with the environment. Moreover, this work introduces GSplat representation which distills latent codes for language semantics and grasp affordance into the 3D scene for scene understanding.

\subsubsection{Navigation}
\label{sec:Navigation}
Navigation in robotics involves two essential and interconnected components: \textit{localization} and \textit{path planning}. Localization deals with the challenge of locating the robot's own position within the environment. Based on the localization result, the robot performs path planning, which refers to the process of determining an optimal route to reach the destination. 3DGS serves as a detailed scene representation for high-precision navigation tasks.

\begin{figure*}[!t]
    \centering \includegraphics[width=\linewidth]{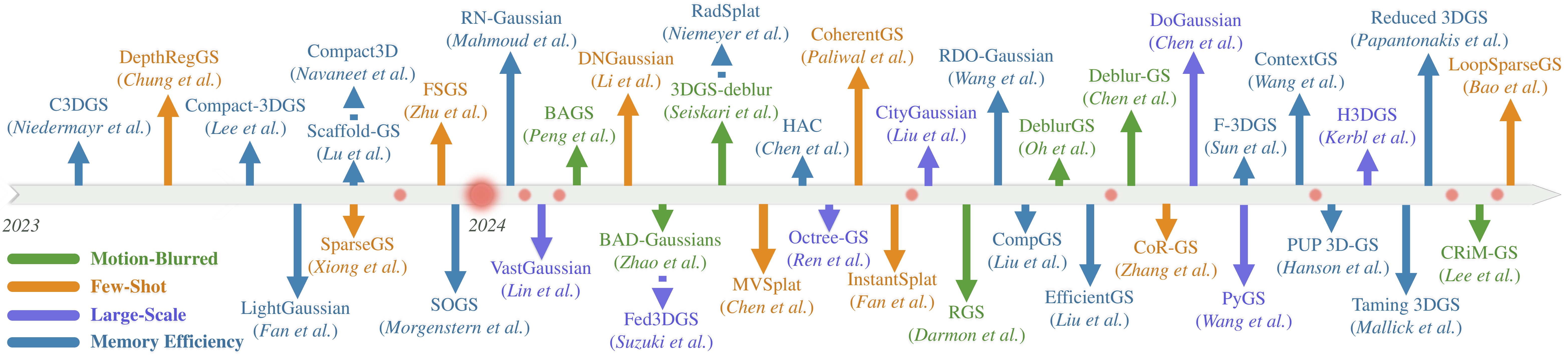}
    \vspace{-0.2in}
    \caption{Chronological: Advance of 3DGS.}
    \label{fig:time_advance}
    \vspace{-0.1in}
\end{figure*}
\begin{figure}[!t]
\centering
    \begin{subfigure}{\linewidth}
    \centering
    \includegraphics[width=\linewidth]{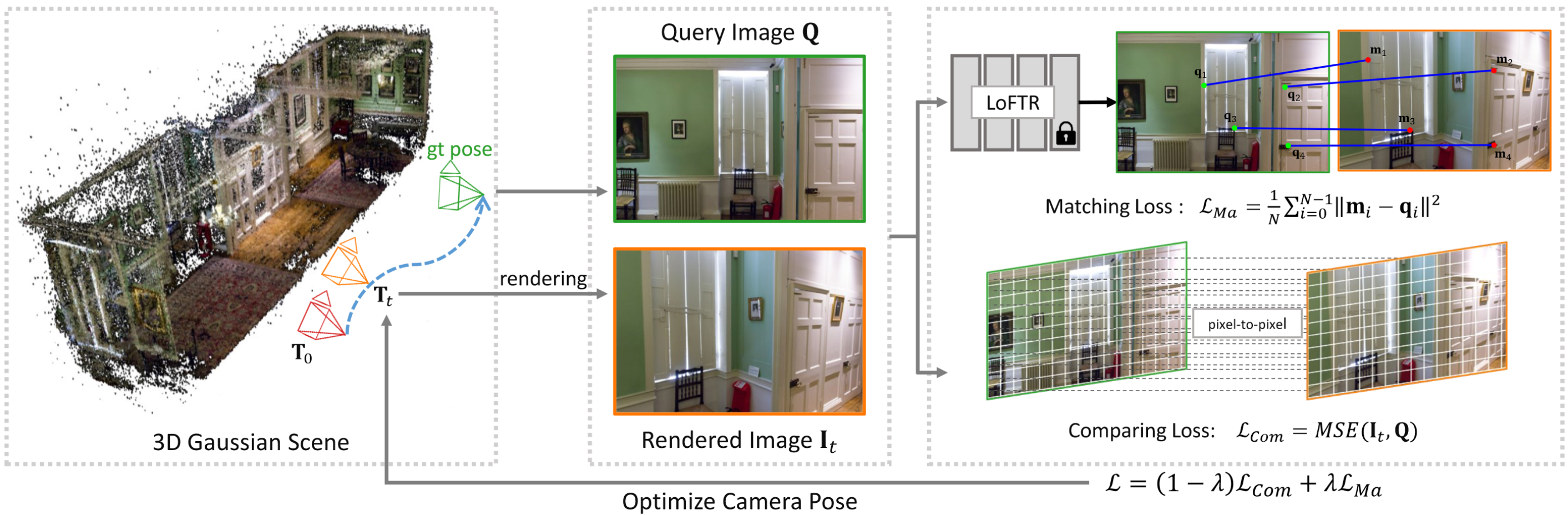}
    \caption{Known map-based localization}
    \vspace{0.05in}
    \label{fig:localization_map_based}
    \end{subfigure}
    
    \begin{subfigure}{\linewidth}
    \centering
    \includegraphics[width=\linewidth]{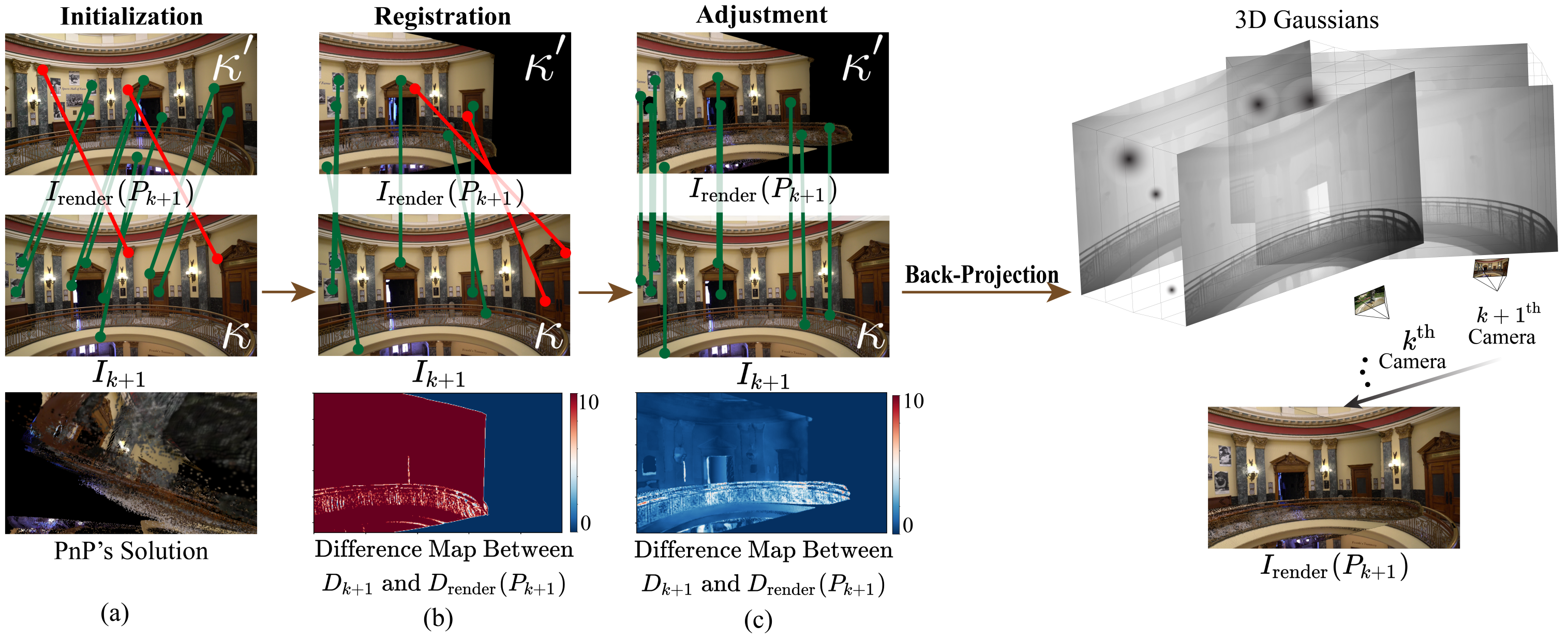}
    \caption{Relative Pose Regression}
    \label{fig:localization_unknown}
    \end{subfigure}
\vspace{-0.15in}
\caption{An illustration of 3DGS for localization. Fig. \ref{fig:localization_map_based} and Fig. \ref{fig:localization_unknown} are originally shown in \cite{sun2023icomma} and \cite{jiang2024construct}, respectively.}
\label{fig:localization}
\vspace{-0.1in}
\end{figure}
\noindent\textbf{Localization.}\hspace{5pt}
Localization refers to estimating a 6 Degree-Of-Freedom (DoF) pose (position and orientation) through the processing of sensor data. We categorize 3DGS-based localization into \textit{known map-based localization} and \textit{relative pose regression} based on whether the prior global map is available, as shown in Fig.~\ref{fig:localization}.

For known map-based localization, iComMa~\cite{sun2023icomma} computes the residuals between the query image and the rendered image obtained from a prebuilt Gaussian map to optimize the camera pose.  
3DGS-ReLoc~\cite{jiang20243dgs} stores 3DGS map as a 2D voxel map with a KD-tree for efficient spatial queries and achieves relocalization by brute-force search to match the query image within the global map. 
Liu {\it et al.}~\cite{liu2024enhancing} generate pseudo scene coordinates from 3DGS for initializing and enhancing scene coordinate regression.
6DGS~\cite{bortolon20246dgs} estimates the camera location by selecting a bundle of rays projected from the ellipsoid surface of Gaussian map and learning an attention map to output ray pixel correspondences for pose optimization. 

In terms of relative pose regression, CF-3DG~\cite{fu2023colmapfree} estimates the relative camera pose that can transform the local 3D Gaussian map of the last frame to render the pixels that align with the current frame.
GGRt~\cite{li2024ggrt} designs a joint learning framework that consists of an iterative pose optimization network that estimates relative poses and a generalizable 3D-Gaussians model that predicts Gaussians.
COGS~\cite{jiang2024construct} detects 2D correspondences between training views and the corresponding rendered images to regress relative pose.
GaussReg~\cite{chang2024gaussreg} explores the registration of 3D scenes using 3DGS representation to estimate the relative pose of the scenes. GSLoc~\cite{liu2024gsloc} incorporates an exposure-adaptive module into the 3DGS model to improve the robustness of matching between two images under domain shift, achieving accurate pose regression of images.

\noindent\textbf{Path planning.}\hspace{5pt}
We categorize path planning into active planning and non-active planning based on whether the robot actively explores the environment for reconstruction as well as planning. 

In terms of active planning, GS-Planner~\cite{jin2024gs} 
maintains a voxel map to represent unobserved volumes. This voxel map is integrated into the splatting process, enabling the exploration of unobserved space. Moreover, this work leverages physical concept of opacity to formulate a chance constraint for safe active planning in 3DGS maps.
For non-active planning, GaussNav~\cite{lei2024gaussnav} constructs a semantic Gaussian map and transfers this map into 2D BEV grids for navigation. Furthermore, a set of descriptive images is generated by rendering the object from multiple viewpoints. These images are then used to match and identify the target position for path planning.
Splat-Nav~\cite{chen2024splat} generates a safe path by discretizing free space on a Gaussian map for safe navigation. This discretization is achieved through intersection tests between Gaussian ellipsoids. 
Beyond Uncertainty~\cite{liu2024beyond} leverages 3DGS map to dynamically assess collision risks at each waypoint and guides risk-aware next-best-view selection for efficient and safe robot navigation.

\begin{figure}[!t]
    \centering 
    \includegraphics[width=\linewidth]{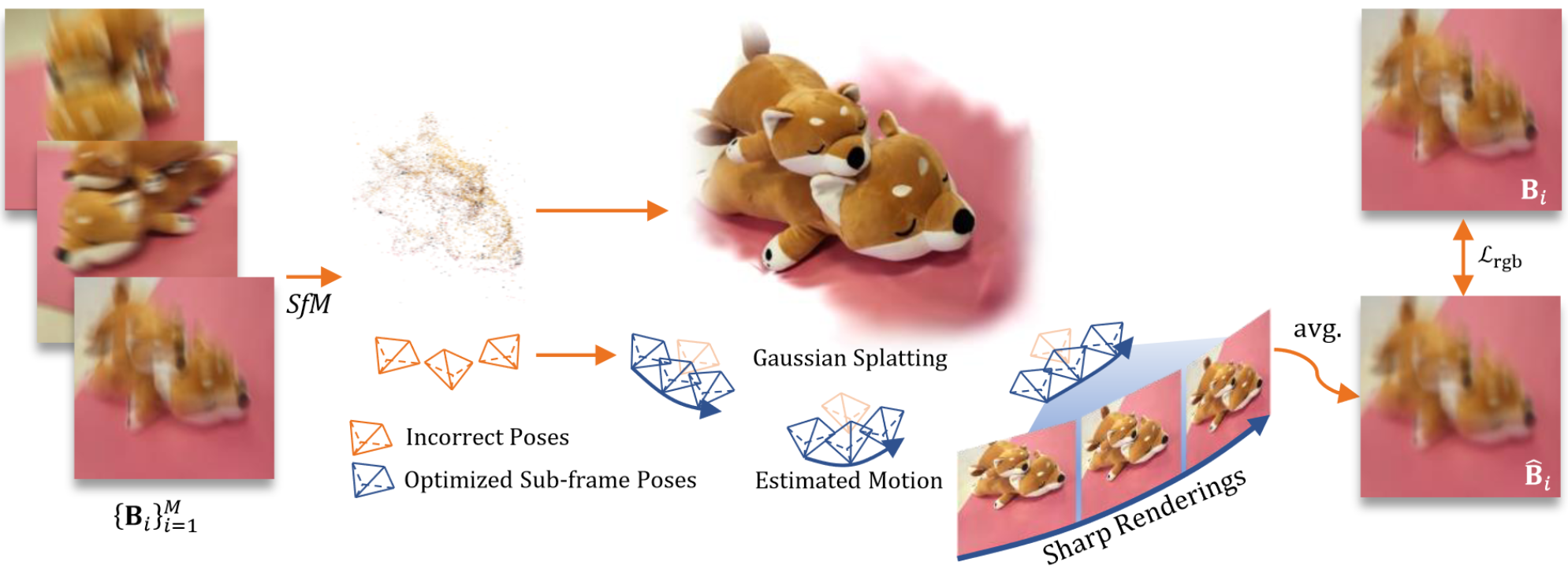}
    \caption{An illustration of 3DGS for deblurring motion blur. Figure is originally shown in \cite{oh2024deblurgs}.}
    \label{fig:deblur}
    \vspace{-0.1in}
\end{figure}
\section{Advance of 3DGS in Robotics}
\label{sec:advance}
While vanilla 3DGS~\cite{kerbl20233d} provides a foundation for 3D scene representation, it has limitations that can impact its effectiveness in robotics applications. To address these, various advanced models have introduced adaptive modifications, improving 3DGS’s adaptability, memory efficiency, and data efficiency for real-world use. Fig.~\ref{fig:time_advance} highlights a timeline of developments that enhance 3DGS specifically for robotic applications, focusing on key improvements to make it more resilient and efficient in complex environments.

\subsection{Adaptability}
\label{sec:Adaptability}
Enhancing the adaptability of vanilla 3DGS involves improving its performance in \textit{large-scale} environment and its resilience to \textit{motion blur}, enabling more effectiveness and reliability across diverse and unpredictable scenarios.
\subsubsection{Motion-Blurred}
\label{sec:Motion blur}
Motion blur of captured images is a common challenge in robotics, primarily caused by high-speed robot movements and slow shutter speed, resulting in degraded image quality. Consequently, deblurring is essential to restore image quality and enhance visual perception for robotic systems. Deblurring methods can be divided into two categories: \textit{physical modeling}, which focuses on understanding and simulating the blurring process based on the formation of motion blur, and \textit{implicit modeling}, which leverages MLP networks to directly learn the mapping between blurred and sharp images without explicitly modeling the underlying physical process, as illustrated in Fig.~\ref{fig:deblur}. 

Physical modeling methods~\cite{zhao2024bad, oh2024deblurgs, chen2024deblur} simulate motion-blurred images by averaging virtual sharp images captured during the camera exposure time. 
These methods then construct loss between simulated and observed blurry images for optimization of Gaussian representation and camera pose, ensuring that the constructed scene is sharp and free from motion blur.
RGS~\cite{darmon2024robust} models motion blur as a Gaussian distribution over camera poses to obtain the expected image at the given noise for optimization. 
3DGS-deblur~\cite{seiskari2024gaussian} simplifies motion blur modeling by approximating motion in pixel coordinates and adjusting the Gaussian means to reflect this movement.
CRiM-GS~\cite{lee2024crim} applies neural ODEs~\cite{chen2018neural} to model the continuous camera motion trajectory during the exposure time and achieves deblurred images by rendering and averaging images from multiple poses sampled along this trajectory.
In terms of implicit modeling, BAGS~\cite{peng2024bags} introduces a blur proposal network to model image blur and produces a quality-assessing mask that indicates regions where blur occur.

\begin{figure}[!t]
\centering
    \begin{subfigure}{\linewidth}
    \centering
    \includegraphics[width=\linewidth]{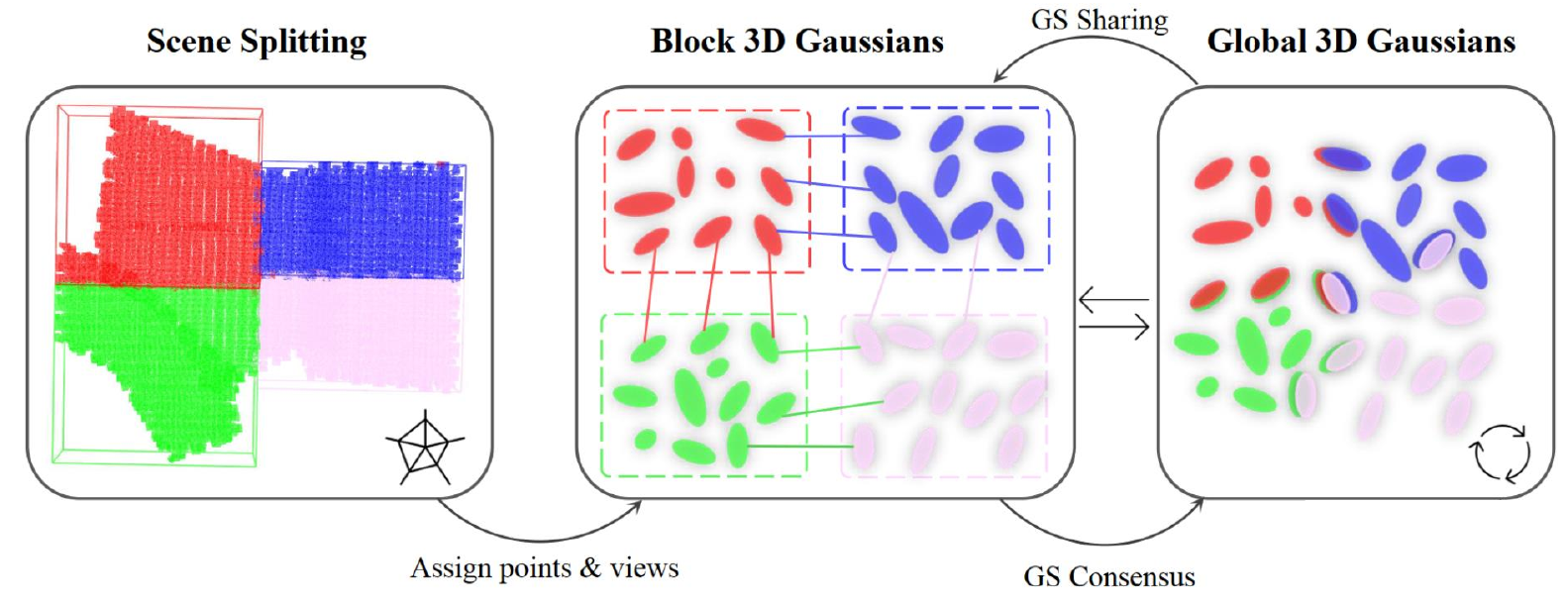}
    \caption{Block separation}
    \vspace{0.05in}
    \label{fig:large_scale_block}
    \end{subfigure}
    
    \begin{subfigure}{\linewidth}
    \centering
    \includegraphics[width=\linewidth]{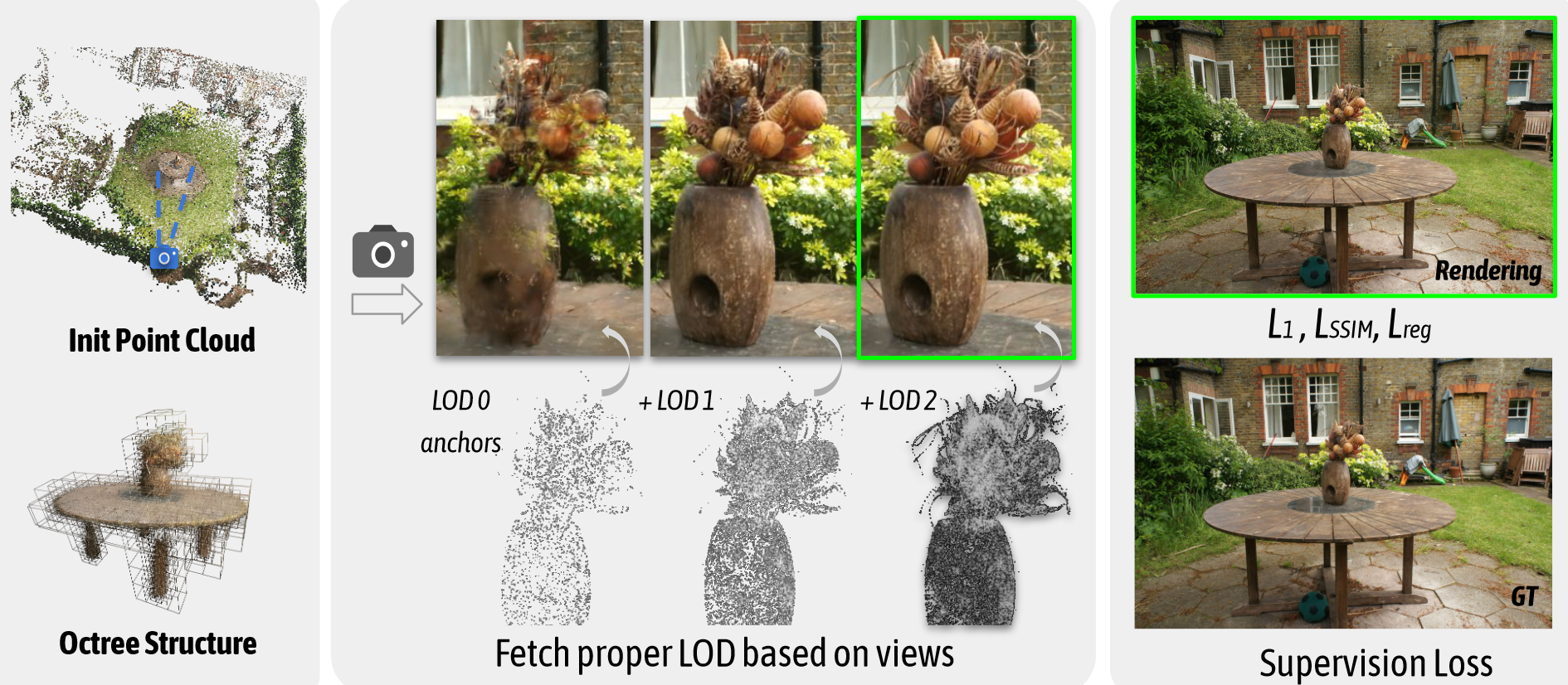}
    \caption{Level-of-Detail}
    \label{fig:large_scale_lod}
    \end{subfigure}
\vspace{-0.2in}
\caption{An illustration of 3DGS for large-scale reconstruction. Fig. \ref{fig:large_scale_block} and Fig. \ref{fig:large_scale_lod} are originally shown in \cite{chen2024dogaussian} and \cite{ren2024octree}, respectively.}
\vspace{-0.05in}
\label{fig:large_scale}
\end{figure}
\subsubsection{Large-Scale}
Vanilla 3DGS representation requires millions of 3D Gaussians for large-scale reconstruction, leading to high GPU memory demands for training, as well as long training and rendering time. To address large-scale reconstruction, existing methods~\cite{liu2024citygaussian, chen2024dogaussian, kerbl2024hierarchical, ren2024octree, suzuki2024fed3dgs, wang2024pygs, lin2024vastgaussian}  divide the scene into independent blocks for separate training (Block separation), and model the scene with hierarchical resolution levels (Level-of-Detail), as shown in Fig.~\ref{fig:large_scale}.

DoGaussian~\cite{chen2024dogaussian}, H3DGS~\cite{kerbl2024hierarchical}, and CityGS~\cite{liu2024citygaussian} employs a divide-and-conquer strategy that partitions the scene into spatially adjacent blocks, which are optimized independently and in parallel. Fed3DGS~\cite{suzuki2024fed3dgs} utilizes decentralized processing of large-scale data across millions of clients to reduce the computational load on a central server.
Octree-GS~\cite{ren2024octree} incorporates an octree structure for multiple-level scene representation and employs level-of-detail decomposition for efficient rendering. 
PyGS~\cite{wang2024pygs} employs a hierarchical structure of 3D Gaussians organized into pyramid levels to represent scenes at varying levels of detail for large-scale scene representation.
VastGaussian~\cite{lin2024vastgaussian} introduces a progressive partitioning strategy to divide a large scene into multiple cells for parallel optimization.

\subsection{Efficiency}
\label{sec:Efficiency}
Vanilla 3DGS suffers from significant storage demands due to its explicit representation and requires extensive multi-view information for scene modeling. 
However, in real-world robotic applications, this substantial storage requirement is inefficient for storing global maps. Additionally, the need to obtain extensive multi-view data for reconstruction leads to data inefficiency. Therefore, improving the efficiency of 3DGS in robotics focuses on two main aspects: \textit{memory efficiency}, which aims to reduce the large storage demands, and \textit{few-shot}, which enhances data utilization by enabling effective reconstruction from minimal multi-view information.

\begin{figure}[!t]
\centering 
\includegraphics[width=\linewidth]{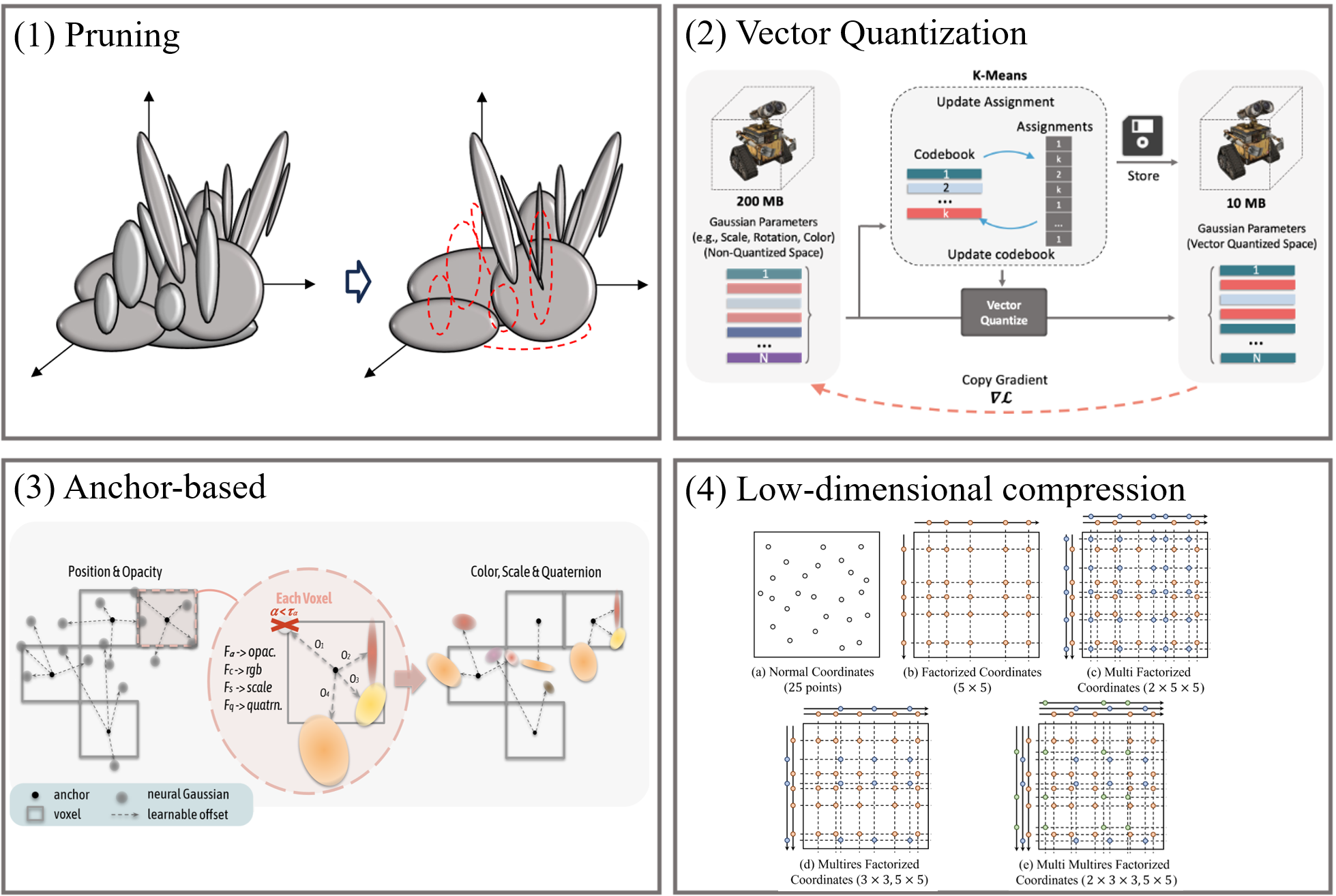}
\vspace{-0.1in}
\caption{An illustration of 3DGS for memery efficiency. Subplot(1) to subplot(4) are extracted from \cite{lee2024compact, navaneet2023compact3d, lu2024scaffold, sun2024f}, respectively.}
\vspace{-0.05in}
\label{fig:memory_efficient}
\end{figure}
\subsubsection{Memery Efficiency}
Explicit representation of 3D Gaussians requires a significant amount of storage space. To address this issue, existing methods primarily adopt four approaches: \textit{pruning}, \textit{vector quantization}, \textit{anchor-based}, and \textit{low-dimensional compression}, to reduce the storage space required for 3D Gaussian scene representation in robotic applications, as shown in Fig.~\ref{fig:memory_efficient}.

Intuitively, reducing the number of 3D Gaussian primitives that contribute little to the scene can decrease the overall storage requirements, which refers to as pruning~\cite{lee2024compact, wang2024end, fan2023lightgaussian, niemeyer2024radsplat, liu2024efficientgs, hanson2024pup, papantonakis2024reducing, mahmoud2024reducing, subhajyoti2024taming}. Compact-3DGS~\cite{lee2024compact} and RDO-Gaussian~\cite{wang2024end} remove Gaussians with small scales and low opacities. LightGaussian~\cite{fan2023lightgaussian}, RadSplat~\cite{niemeyer2024radsplat}, and EfficientGS~\cite{liu2024efficientgs} prune Gaussians that have low ray-contribution to the pixels across all training views. PUP 3D-GS~\cite{hanson2024pup} employs a sensitivity score derived from Hessian matrix of reconstruction errors, pruning Gaussians with higher spatial uncertainty and lesser contributions to reconstruction quality.
Reduced 3DGS~\cite{papantonakis2024reducing} identifies regions that are densely populated with Gaussian primitives and prunes Gaussians that overlap significantly with other Gaussians. RN-Gaussian~\cite{mahmoud2024reducing} and Taming 3DGS~\cite{subhajyoti2024taming} restrict the cloning and splitting of Gaussians based on position and correlation with existing Gaussians to reduce the number of Gaussian.
Moreover, Vector Quantization methods~\cite{niedermayr2024compressed, navaneet2023compact3d, lee2024compact, papantonakis2024reducing} convert Gaussian parameters into codebooks to represent Gaussian attributes compactly. 

Anchor-based methods~\cite{lu2024scaffold, chen2024hac, liu2024compgs, wang2024contextgs} employ a set of anchor primitives to cluster related nearby 3D Gaussians and predict their attributes from anchors’ attributes, resulting in memory-efficient 3D scene representation. HAC~\cite{chen2024hac} leverages relations between the unorganized anchors and the structured hash grid for context modeling of anchor attributes. ContextGS~\cite{wang2024contextgs} divides anchors into multiple levels for compact representation. Moreover, low-dimensional compression methods~\cite{morgenstern2023compact, sun2024f} represent 3D Gaussian in 1D or 2D space for compact storage. SOGS~\cite{morgenstern2023compact} maps originally unstructured 3D Gaussians to 2D grids for reducing memory. F-3DGS~\cite{sun2024f} proposes a factorized coordinate scheme that maintains 1D or 2D coordinates in each axis or plane and generates 3D coordinates by a tensor product.

\subsubsection{Few-Shot}
Novel view rendering from few-shot reconstruction is challenging due to the sparse information available. In scenarios where only limited observations are present, vanilla 3DGS faces two challenges: \textit{(1)} Sparse SfM points derived from few-shot images fail to represent the global geometry for Gaussian initialization. \textit{(2)} Leveraging a few images for optimization can lead to overfitting. To address these challenges, few-shot methods~\cite{zhu2023fsgs, chung2024depth, paliwal2024coherentgs, chen2024mvsplat, xiong2023sparsegs, bao2024loopsparsegs, zhang2024cor, li2024dngaussian, fan2024instantsplat} incorporates additional constraints and prior into SfM initialization and Gaussian optimization.

FSGS~\cite{zhu2023fsgs} inserts new Gaussians near existing ones based on their proximity to deal with sparse initial point sets and generates unseen viewpoints for training to address overfitting. 
DepthRegGS~\cite{chung2024depth} employs depth-guided optimization to mitigate overfitting. 
CoherentGS~\cite{paliwal2024coherentgs} proposes depth-based initialization to achieve dense SfM points.
MVSplat~\cite{chen2024mvsplat} extracts multi-view image features to construct per-view cost volumes and predicts per-view depth maps for few-shot reconstruction. 
SparseGS~\cite{xiong2023sparsegs} integrates depth priors with generative and explicit constraints to enhance consistency from unseen viewpoints.
LoopSparseGS~\cite{bao2024loopsparsegs} incorporates iteratively rendered images with training images into SfM to densify the initialized point cloud. Additionally, this work leverages window-based dense monocular depth to provide precise geometric supervision for Gaussian optimization.

\section{Datasets and Performance Evaluation}
\label{sec:performance}
In this section, we present a detailed summary of the
datasets commonly used in robotics. Moreover, we provide performance comparison and evaluation of key modules in robotics, including mapping and localization, perception, manipulation and navigation. These modules are fundamental and crucial to the functionality of intelligent robotic systems. By comparing the performance of these modules, we aim to offer an overview of the current state of research in robotics based on 3DGS representation. In terms of perception, we only report on scene segmentation, as there are currently no other 3DGS-based perception methods available at the time of this paper's completion.

\subsection{SLAM Evaluation}
\noindent\textbf{Datasets.}\hspace{5pt}
Existing 3DGS-based visual SLAM and semantic SLAM methods use indoor scene datasets for evaluation, including Replica~\cite{straub2019replica}, ScanNet~\cite{dai2017scannet}, and TUM RGB-D~\cite{sturm2012benchmark}. As TUM RGB-D dataset~\cite{sturm2012benchmark} lacks semantic annotations, it is generally not used for benchmarking semantic SLAM methods. For multi-sensor fusion SLAM methods, evaluations are conducted on different outdoor scene datasets based on the specific sensors used, rather than a unified dataset. Detailed descriptions of the SLAM datasets are provided below.

\begin{itemize}
    \item \textit{Replica}~\cite{straub2019replica} is an RGB-D dataset that contains 18 indoor scene reconstructions at room and building scales. Each scene consists of dense mesh, high-dynamic-range (HDR) textures, per-primitive semantic class and instance information.
    \item \textit{ScanNet dataset}~\cite{dai2017scannet} contains RGB-D frames and IMU data of 2.5M views in 1513 scenes, each annotated with 3D camera poses, surface reconstructions, textured meshes, and semantic segmentations.
    \item \textit{TUM RGB-D}~\cite{sturm2012benchmark} is an RGB-D dataset that consists of 39 sequences recorded in an office environment and an industrial hall. RGB-D data are captured from the Kinect. Ground truth pose estimates are obtained from the motion capture system.
\end{itemize}

\noindent\textbf{Metrics.}\hspace{5pt}
Reconstruction and tracking metrics are employed for evaluating SLAM accuracy. Specifically, \textit{Depth L1 (cm)} metric is used for reconstruction evaluation, which is the average absolute error between ground truth depth and reconstructed depth. \textit{ATE RMSE (cm)}~\cite{sturm2012benchmark} is tracking metric that quantifies the error between the estimated trajectory and the ground truth trajectory. For SLAM systems, real-time performance is crucial, and metric \textit{FPS} is utilized for evaluating the time consumption of SLAM process. In addition, \textit{mIoU (\%)} is used for semantic segmentation evaluation in semantic SLAM that measures the average percentage of overlap between the predicted and ground truth areas across different classes. Moreover, as most 3DGS-based SLAM methods report the rendering performance of SLAM mapping results, we also include rendering metrics in the SLAM evaluation. Peak signal-to-noise ratio \textit{(PSNR[dB])}, structural similarity \textit{(SSIM)}~\cite{wang2004image}, and learned perceptual image patch similarity \textit{(LPIPS)}~\cite{zhang2018unreasonable} are rendering metrics.

\noindent\textbf{Results.}\hspace{5pt}
As visual SLAM and semantic SLAM methods are mostly tested on Replica dataset~\cite{straub2019replica}, we present performance comparison on Replica regarding SLAM metrics, as shown in Table~\ref{tab:SLAM evaluation}. In addition, time consumption results reported in 3DGS SLAM works are obtained from experiments conducted on different GPUs. Since the computational capabilities of GPUs vary significantly, it is unfair to directly compare the real-time performance of SLAM methods based on these results. Therefore, FPS metric is not reported.

    
\begin{table}[t]
    \caption{SLAM performance comparison. The results are average of 8 scenes on Replica dataset~\cite{straub2019replica}. Best results are highlighted as \colorbox{NO1}{first}, \colorbox{NO2}{second}. These notes also apply to the other tables.}
    \vspace{-0.05in}
    \label{tab:SLAM evaluation}
    \centering
    \resizebox{\linewidth}{!}{
    \begin{tabular}{cc|l|c|ccc|c|c}
    \toprule
     & & \multirow{2}{*}{Methods} & Reconstruction & \multicolumn{3}{c|}{Rendering} & Tracking & Segmentation \\
    \cline{4-9}
     & & & Depth L1$\downarrow$ &	PSNR$\uparrow$	& SSIM$\uparrow$ &	LPIPS$\downarrow$ &	RMSE$\downarrow$ &	mIoU$\uparrow$ \\
     \cline{1-9}
    \multirow{17}{*}{\rotatebox{90}{Visual SLAM}} & \multirow{12}{*}{\rotatebox{90}{RGB-D}} & GS-SLAM~\cite{yan2024gs} & 1.16 & 34.27 & 0.98 & 0.08 & 0.50 &  -- \\
    & & Photo-SLAM~\cite{huang2024photo}  & -- & 34.96 & 0.94 & 0.06  & 0.60 &  -- \\
    & & SplaTAM~\cite{keetha2024splatam} & 0.72 & 34.11 & 0.97 & 0.10 & 0.36 & -- \\
    & & Gaussian-SLAM~\cite{yugay2023gaussian} & 1.56 & \cellcolor{NO2}38.90 & \cellcolor{NO1}0.99 & 0.07 & 0.48 &  -- \\
    & & MonoGS~\cite{matsuki2024gaussian} & -- & 37.50 & 0.96 & 0.07 & 0.32 &  -- \\
    & & Compact-SLAM~\cite{deng2024compact} & -- & 34.44 & 0.98 & 0.09 & 0.33 & -- \\
    & & GS-ICP SLAM~\cite{ha2024rgbd} & -- & 38.83 & 0.98 & \cellcolor{NO1}0.04 & \cellcolor{NO1}0.16 &  -- \\
    & & HF-SLAM~\cite{sun2024high} & \cellcolor{NO2}0.52 & 36.19 & 0.98 & \cellcolor{NO2}0.05 & 0.25 &  -- \\
    & & NGM-SLAM~\cite{li2024ngm} & -- & 37.43 & 0.98 & 0.08 & 0.51 &  -- \\
    & & MotionGS~\cite{guo2024motiongs} & -- & \cellcolor{NO1}39.60 & 0.98 & \cellcolor{NO1}0.04 & 0.49 &  -- \\
    & & RTG-SLAM~\cite{peng2024rtg} & -- & 35.43 & 0.98 & 0.11 & \cellcolor{NO2}0.18 &  -- \\
    & & CG-SLAM~\cite{hu2024cg} & -- & -- & -- & --  & 0.27 & -- \\
    & & LoopSplat~\cite{zhu2024loopsplat} & \cellcolor{NO1}0.51 & 36.63 & \cellcolor{NO1}0.99 & 0.11 & 0.26 & -- \\
    \cline{2-9}
    & \multirow{5}{*}{\rotatebox{90}{RGB}} & Photo-SLAM(RGB)~\cite{huang2024photo} & -- & 33.30 & 0.93 & 0.08 &  1.09 & --\\
    & & NGM-SLAM(RGB)~\cite{li2024ngm} & -- & 35.02 & \cellcolor{NO1}0.96 & 0.13 & 8.51 & --\\
    & & MGS-SLAM~\cite{zhu2024mgs} & -- & 29.90 & 0.88 & 0.09 & \cellcolor{NO1}0.32 & --\\
    & & MonoG-SLAM~\cite{lan2024monocular} & -- & 33.59 & 0.93 & 0.22 & \cellcolor{NO1}0.32 & --\\
    & & Splat-SLAM~\cite{sandstrom2024splat} & \cellcolor{NO1}2.41 & \cellcolor{NO1}36.45 & \cellcolor{NO2}0.95 & \cellcolor{NO2}0.06 & \cellcolor{NO2}0.34 & --  \\
    & & IG-SLAM~\cite{sarikamis2024ig} & \cellcolor{NO2}4.33 & \cellcolor{NO2}36.21 & \cellcolor{NO1}0.96 & \cellcolor{NO1}0.05 & 0.44 & -- \\
    \hline
    \multicolumn{2}{c|}{\multirow{4}{*}{\rotatebox{90}{\begin{tabular}{@{}c@{}}Semantic \\ SLAM\end{tabular}}}}
    & SGS-SLAM~\cite{li2024sgs} & \cellcolor{NO1}0.36 & 34.66 & 0.97 & 0.10 & 0.41 & 92.72\\ 
    & & SemGauss-SLAM~\cite{zhu2024semgauss}& 0.50 & \cellcolor{NO2}35.03 & \cellcolor{NO2}0.98 & \cellcolor{NO2}0.06 & \cellcolor{NO1}0.33 & \cellcolor{NO2}96.34\\
    & & NEDS-SLAM~\cite{ji2024neds}& \cellcolor{NO2}0.47 & 34.76 & 0.96 & 0.09 & \cellcolor{NO2}0.35 & 90.78\\ 
    & & GS$^3$LAM~\cite{li2024gs} & -- & \cellcolor{NO1}36.26 & \cellcolor{NO1}0.99 & \cellcolor{NO1}0.05 & 0.37 & \cellcolor{NO1}96.63\\
    \toprule
    \end{tabular}
    }
\end{table}

\subsection{Scene Reconstruction Evaluation}
\noindent\textbf{Datasets.}\hspace{5pt}
Indoor reconstruction typically utilizes datasets similar to those used in SLAM. For outdoor scene reconstruction, commonly used datasets include KITTI~\cite{geiger2012we}, Waymo~\cite{sun2020scalability}, nuScenes~\cite{caesar2020nuscenes}, and Argoverse~\cite{chang2019argoverse} dataset. 
Detailed descriptions of outdoor reconstruction datasets are shown below.
\begin{itemize}
    \item \textit{KITTI dataset}~\cite{geiger2012we} contains data from 4 cameras and 1 LiDAR, consisting of 22 stereo sequences with a total length of 39.2km. This dataset also comprises 200k 3D object annotations and 389 optical flow image pairs.
    \item \textit{Waymo dataset}~\cite{sun2020scalability} contains data from 5 cameras and 5 LiDAR sensors, consisting of 1150 scenes across a range of urban and suburban geographies. This dataset is annotated with 2D (camera image) and 3D (LiDAR) bounding boxes.
    \item \textit{nuScenes dataset}~\cite{caesar2020nuscenes} contains data from 6 cameras, 5 radars, 1 lidar sensor, and IMU, comprising 1000 scenes, each fully annotated with 3D bounding boxes for 23 classes and 8 attributes. 
    \item \textit{Argoverse dataset}~\cite{chang2019argoverse} contains sequences of LiDAR measurements, 360$^\circ$ images from 7 cameras, and forward-facing stereo imagery. This dataset also includes annotations for 290km of mapped lanes and 300k interesting vehicle trajectories.
\end{itemize}

\begin{table}[t]
    \caption{Dynamic scene reconstruction performance comparison. The results are novel view rendering performance on KITTI dataset~\cite{geiger2012we} using 75\% and 50\% of full data for training respectively. }
    \vspace{-0.05in}
    \label{tab:Reconstruction scene evaluation}
    \centering
    \resizebox{\linewidth}{!}{
    \begin{tabular}{l|ccc|ccc}
    \toprule
    \multirow{2}{*}{Methods} & \multicolumn{3}{c}{KITTI [75\%]} & \multicolumn{3}{c}{KITTI [50\%]} \\
    \cline{2-4}
    \cline{5-7}
    & PSNR$\uparrow$	& SSIM$\uparrow$ & LPIPS$\downarrow$ &	PSNR [dB]$\uparrow$	& SSIM$\uparrow$ &	LPIPS$\downarrow$	\\
    \hline
    PVG~\cite{chen2023periodic}	& \cellcolor{NO2}27.43 & 0.896 & 0.114 & -- & -- & -- \\
    VDG~\cite{li2024vdg}  &	25.29 &	0.851 &	0.152 & -- & -- & -- \\
    Street Gaussians~\cite{yan2024street}	 &	25.79	&	0.844	&	\cellcolor{NO2}0.081	& 25.52	&	0.841	&	\cellcolor{NO2}0.084	\\
    AutoSplat~\cite{khan2024autosplat} & 26.59 & \cellcolor{NO2}0.913 & 0.204 & \cellcolor{NO2}26.22 &	\cellcolor{NO2}0.907 & 0.207 \\
    4DGF~\cite{fischer2024dynamic} &\cellcolor{NO1}31.34 & \cellcolor{NO1}0.945 & \cellcolor{NO1}0.026 & \cellcolor{NO1}30.55 & \cellcolor{NO1}0.931 & \cellcolor{NO1}0.028 \\
    \toprule
    \end{tabular}}
\end{table}

\noindent\textbf{Metrics.}\hspace{5pt}
Rendering metrics \textit{PSNR}, \textit{SSIM}, and \textit{LPIPS} are employed for evaluation.

\noindent\textbf{Results.}\hspace{5pt}
In terms of static reconstruction, various methods use different scenes from these datasets for evaluation rather than using specific scenes. For dynamic scene reconstruction, most methods employ consistent training and testing splits on the KITTI dataset~\cite{geiger2012we} for evaluation. We present performance comparison on KITTI~\cite{geiger2012we} regarding rendering metrics, as shown in Table~\ref{tab:Reconstruction scene evaluation}. Some dynamic scene reconstruction methods that do not report KITTI results are not displayed in the table.

\subsection{Localization Evaluation}
\noindent\textbf{Datasets.}\hspace{5pt}
Evaluation of localization performance typically uses Tanks\&Temples~\cite{knapitsch2017tanks}, 7 Scenes~\cite{shotton2013scene} datasets, and autonomous driving datasets that are similar to those used in outdoor scene reconstruction. Detailed descriptions of localization datasets are as follows.
\begin{itemize}
    \item \textit{Tanks\&Temples dataset}~\cite{knapitsch2017tanks} contains high-resolution video sequences and ground truth poses of both indoor and outdoor scenes.
    \item \textit{7 Scenes}~\cite{shotton2013scene} is an RGB-D dataset with ground truth poses and dense 3D models. The sequences were recorded for each scene by different users and split into distinct training and testing sets for localization evaluation.
\end{itemize}

\noindent\textbf{Metrics.}\hspace{5pt}
Pose metrics \textit{Absolute Trajectory Error (ATE)} and \textit{Relative Pose Error (RPE)} are used for localization evaluation. ATE quantifies the difference between the estimated camera positions and the ground truth positions. RPE measures the relative pose errors between pairs of images. In 3DGS-based localization, rendering metrics \textit{PSNR}, \textit{SSIM}, and \textit{LPIPS} are also employed for evaluating the accuracy of images rendered based on the estimated poses.

\noindent\textbf{Results.}\hspace{5pt}
Existing 3DGS-based localization methods employ various datasets and sequences. Meanwhile, these methods are not open source, so it is not possible to compare the performance of existing localization methods under the same conditions. Generally, existing 3DGS-based localization methods can achieve a positional accuracy of 5$cm$ and an angular accuracy of 2$^\circ$.


\subsection{Segmentation Evaluation}
\noindent\textbf{Datasets.}\hspace{5pt}
Datasets with ground truth semantic annotations are used for segmentation evaluation, including LERF-Mask~\cite{ye2023gaussian}, SPIn-NeRF~\cite{mirzaei2023spin}, as well as indoor datasets Replica~\cite{straub2019replica} and ScanNet~\cite{dai2017scannet}, which are also used in SLAM evaluations.  Specifically, LERF-Mask~\cite{ye2023gaussian} 
is used to assess text-query semantic segmentation, a crucial task in robotic manipulation and navigation that requires robots to identify specific objects in the environment following text prompts. Detailed descriptions of segmentation datasets are given below.
\begin{itemize}
    \item \textit{LERF-Mask dataset}~\cite{ye2023gaussian} contains semantic annotations of three scenes from LERF-Localization dataset~\cite{kerr2023lerf}. This dataset contains a total of 23 prompts.
    \item \textit{SPIn-NeRF dataset}~\cite{mirzaei2023spin} contains 10 real-world forward-facing scenes with annotated object masks. Each scene includes 60 training images with the object and 40 test images without the object.
\end{itemize}

\noindent\textbf{Metrics.}\hspace{5pt}
Although existing 3DGS-based semantic segmentation methods can achieve 3D segmentation, their performance is still evaluated by measuring the discrepancy between the rendered semantics and the 2D ground truth labels.
Mean intersection over union \textit{(mIoU(\%))}, mean boundary intersection over union \textit{(mBIoU(\%))}, and pixel accuracy \textit{(Acc.(\%))} are used for semantic evaluation. Specifically, mIoU measures the overlap of the ground truth and rendered masks. mBIoU quantifies contour alignment between predicted and ground truth masks.

\begin{table}[t]
    \caption{Segmentation performance comparison.}
    \vspace{-0.05in}
    \label{tab:Segmentation evaluation}
    \scriptsize
    \centering
    \begin{tabular}{l|c|cc}
    \toprule
    \multirow{2}{*}{Methods} & LERF-Mask & \multicolumn{2}{c}{SPIn-NeRF}\\
    \cline{2-4}
    & mIoU$\uparrow$ & mIoU$\uparrow$ & Acc.$\uparrow$\\
    \hline
    SAGA~\cite{cen2023saga} &  -- & 88.0 & 98.5 \\
    SAGD~\cite{hu2024semantic} &  -- & 89.9 & \cellcolor{NO1}98.7 \\
    Gaussian Grouping~\cite{ye2023gaussian} &  72.8 & -- & -- \\
    Feature 3DGS~\cite{zhou2024feature} &  65.6 & -- & -- \\
    Gaga~\cite{lyu2024gaga} &  74.7 & -- & -- \\
    CGC~\cite{silva2024contrastive} &  80.3 & -- & -- \\
    Click-Gaussian~\cite{choi2024click} &   \cellcolor{NO1}89.1 &  \cellcolor{NO1}94.0 & -- \\
    \toprule
    \end{tabular}
    \vspace{-0.7mm}
\end{table}

\noindent\textbf{Results.}\hspace{5pt}
Performance comparison on LERF-Mask~\cite{ye2023gaussian} and SPIn-NeRF~\cite{mirzaei2023spin} datasets are shown in Table~\ref{tab:Segmentation evaluation}. 

\subsection{Manipulation Evaluation}
\noindent\textbf{Datasets.}\hspace{5pt}
The performance of manipulation methods is evaluated on both simulated datasets and real-world scenes. Simulated datasets include RLBench~\cite{james2020rlbench} and Robomimic~\cite{mandlekar2021matters}. In addition, real-world robotic arm trials are conducted to measure the grasping accuracy in practical scenarios. Detailed descriptions of the simulated datasets are shown below.
\begin{itemize}
    \item \textit{RLBench dataset}~\cite{james2020rlbench} is a large-scale learning environment featuring 100 unique, hand-designed tasks. These tasks range from simple target reaching and door opening, to longer multi-stage tasks, such as opening an oven and placing a tray in it.
    \item \textit{Robomimic dataset}~\cite{mandlekar2021matters} consists of data collected from three different sources, including Machine-Generated (MG), Proficient-Human (PH), and Multi-Human (MH). MG contains 300 rollout trajectories from agent checkpoints that are saved regularly during training of a reinforcement learning (RL) method~\cite{haarnoja2018soft}. PH consists of 200 demonstrations collected from a single proficient human operator. MH consists of 300 demonstrations collected from six human operators of varied proficiency.
\end{itemize}

\noindent\textbf{Metrics.}\hspace{5pt}
\textit{Success rate (\%)} metric is used to evaluate the performance of manipulation, representing the percentage of successful executions of grasping or placement tasks by a robot arm over multiple tries.

\noindent\textbf{Results.}\hspace{5pt}
In existing 3DGS-based manipulation methods, grippers are employed to perform operations. Due to variations in real-world test scenarios and differences in the robot arms used, it is not feasible to conduct comparisons under the same conditions. Generally, for single-stage tasks, the success rate can reach up to 80\%, whereas for multi-stage tasks, the success rate is under 50\%.

\subsection{Path Planning Evaluation}
\noindent\textbf{Datasets.}\hspace{5pt}
3DGS-based path planning evaluation uses simulated dataset Matterport3D~\cite{chang2017matterport3d}, Habitat-Matterport 3D (HM3D)~\cite{yadav2023habitat} or custom-built simulation environments via the Unity engine. Detailed descriptions of path planning datasets are given below.
\begin{itemize}
    \item \textit{Matterport3D}~\cite{chang2017matterport3d} is a large-scale RGB-D dataset containing 10,800 panoramic views from 90 building-scale scenes, with surface reconstructions, camera poses, 2D and 3D semantic segmentation annotations. This dataset is typically constructed with the Habitat simulator~\cite{savva2019habitat} for evaluating path planning.
    \item \textit{HM3D}~\cite{yadav2023habitat}, which is designed for the object goal navigation task, consists of 142,646 object instance annotations across 216 3D spaces and 3,100 rooms within those spaces.
    
\end{itemize}
\noindent\textbf{Metrics.}\hspace{5pt}
Planning accuracy metric \textit{success rate (\%)} and \textit{SPL}~\cite{anderson2018evaluation}, planning safety metric \textit{Wasserstein distance ($W_2(P,\hat{P})$)}~\cite{givens1984class}, are used to evaluate path planning performance. 
Specifically, success rate is the percentage of trials in which the agent successfully invokes the STOP action within a predefined Euclidean distance from the goal object.
SPL is success rate weighted by normalized inverse path length that considers both the success of reaching the goal and the efficiency (path length) of getting there. 
$W_2(P,\hat{P})$ quantifies the dissimilarity between the robot's estimated risk distribution $\hat{P}$ and the true risk distribution $P$, which indicates the accuracy of the robot's risk assessment of the environment that represents the safety of robots.

\noindent\textbf{Results.}\hspace{5pt}
Existing 3DGS-based path planning methods are designed for various tasks, such as Instance ImageGoal Navigation (IIN) and safe navigation. Therefore, these 3DGS-based methods can only be compared with corresponding traditional navigation methods under the same conditions. Typically, for IIN task, existing methods reach a 72\% success rate of path planning. In save navigation, existing works can achieve an average of 0.68 in $W_2(P,\hat{P})$ metric.



\section{Future Research}
\label{sec:future}
Although 3DGS has been widely used in robotics tasks, there are still many challenges that remain unsolved in such tasks. In this section, we present some valuable research directions as references for future research.

\subsection{Robust Tracking}
Existing 3DGS-based SLAM methods, although demonstrating high accuracy in dense mapping, typically fail to achieve accurate and robust tracking, especially in complex real-world scenarios. This limitation in current 3DGS-based SLAM systems is due to their reliance on directly using RGB information of image for pose optimization. Such reliance heavily depends on the quality and texture information of the images. However, in real-world robotic applications, image quality is prone to camera motion blur, degrading the performance of 3DGS-based SLAM. Moreover, there are some scenes with limited texture information, such as sky or walls, leading to insufficient constraints for pose estimation. The following presents corresponding directions to improve the robustness of tracking.

\noindent\textbf{Camera motion blur.}\hspace{5pt}
Camera motion blur is primarily caused by rapid movements of the robot and slow shutter speed of the camera, leading to blurry images. Although deblurring  methods have been researched (Section \ref{sec:Motion blur}) and used in SLAM~\cite{bae20242}, these methods fail to directly convert captured blurry images into sharp ones. Instead, they simulate motion blur by averaging virtual sharp images captured during the camera exposure time to synthesize blurry images. These synthesized blurry images are then used to construct loss with the observed blurry images for Gaussian optimization, ensuring that the constructed scene is deblurred. However, such methods fail to address the degradation of image quality in the observed images caused by motion blur, which adversely affects tracking performance that relies on high-quality images for pose optimization. A suitable research direction is to leverage the advantages of 3DGS representation, such as geometric information and spatial distribution, to perform tracking. This method can reduce the reliance on image quality.

\noindent\textbf{Limited texture information.}\hspace{5pt}
In real-world scenes, there are some corner cases where the environmental texture information is limited, leading to insufficient constraints for pose optimization that solely relies on image quality. Although some 3DGS-based SLAM methods~\cite{lang2024gaussian, hong2024liv} have utilized multi-sensor fusion traditional SLAM as odometry for tracking, these methods fail when traditional SLAM is unable to handle complex corner cases. A potential research direction is to incorporate original sensor data of multiple sensors, such as IMU, wheel encoders, and LiDAR, with 3D Gaussian representation to provide sufficient constraints for pose optimization. This approach not only leverages the spatial structural information and dense scene representation offered by 3DGS, but also exploits the various constraints from multi-sensor information. 

\subsection{Lifelong Mapping and Localization}
Current 3DGS methods primarily focus on short-term reconstruction and localization. However, in most real-world scenarios, the environment undergoes constant changes over time. A prebuilt map that fails to consider these changes may quickly become outdated and unreliable. Consequently, it is crucial to maintain an up-to-date model of the environment to facilitate the long-term operation or navigation of robots.  
Although some traditional methods~\cite{adkins2024obvi, zhao2021general} have achieved long-term mapping, these approaches focus on constructing and updating sparse maps, which are insufficient for downstream robotic tasks. Therefore, a promising research direction is lifelong 3DGS-based dense mapping and localization. Since 3DGS is an explicit and dense representation, the dynamic update and refinement of the Gaussian map can be achieved through explicit editing of Gaussian primitives. Additionally, we believe that the inconsistencies in the Gaussian map caused by long-term dynamic changes can be optimized by leveraging the inner constraints between Gaussian primitives. Therefore, by harnessing the explicit representation and inherent constraints among Gaussian primitives, lifelong mapping and localization can be achieved.

\subsection{Large-scale Relocalization}
In robotic applications, it is necessary for robots to relocate their current poses upon entering a pre-established map. However, existing 3DGS-based relocalization methods~\cite{liu2024enhancing ,jiang20243dgs} either require a coarse initial pose or are only capable of achieving relocalization in small indoor scenes. These methods struggle to perform relocalization in large-scale outdoor scenes without an initial pose. Unfortunately, it is challenging to obtain a coarse initial pose for relocalization in practical robotic applications. Therefore, a meaningful research direction is large-scale relocalization without prior poses. We believe that constructing a submap index library or descriptor based on 3DGS representation facilitates coarse pose regression. In addition, the coarse pose can be refined through a registration process that leverages geometric and appearance features within the 3DGS representation.

\subsection{Sim-to-Real Manipulation}
Collecting real-world manipulation datasets is challenging, leading to a scarcity of data for training effective grasping in real scenarios. Therefore, grasping methods often require initial training in simulation environments before being transferred to real-world settings. 
Although 3DGS-based sim-to-real method~\cite{wu2024rl} has been explored, it has limitation in generalization. Specifically, this approach heavily depends on scene-specific training, which hinders its ability to generalize and transfer learned knowledge between similar task scenarios. Consequently, this method still requires a substantial amount of real-world datasets for training. 
Furthermore, the discrepancies in material and physical properties between simulation and reality environments can lead to significant differences in training data distributions for manipulation tasks. These discrepancies may potentially result in entirely different operation strategies. However, existing method~\cite{zhang2024physdreamer} only enables modeling the physical properties of real-world scenarios. Therefore, a  promising research direction involves directly incorporating uncertainty and environmental features into the 3DGS representation to enhance generalization and property modeling.

\section{Conclusion}
\label{sec:conclusion}
As a powerful radiance field for dense scene representation, 3DGS provides new options in the field of robotics for scene understanding and interaction. Specifically, 3DGS offers a dependable selection for many applications in robotics, such as reconstruction, scene segmentation, scene editing, SLAM, manipulation, and navigation. Moreover, the capability of 3DGS to enhance its performance in large-scale environments, motion-blurred conditions, few-shot scenarios, etc., remains mostly unexploited. Exploring these areas could significantly deepen the integration between 3DGS and robotics. In addition, we conduct a thorough performance evaluation of current 3DGS applications in robotics, helping readers choose their preferred approach. Finally, we discuss in detail the challenges and future development directions of 3DGS in robotics. Therefore, as our survey provides a comprehensive summary of the field's outstanding work and emphasizes its potential, we hope this survey encourages more researchers to explore new possibilities and successfully implement them on real robotic platforms.


\ifCLASSOPTIONcompsoc
\else
  \section*{Acknowledgment}
\fi

\bibliographystyle{IEEEtran}
\bibliography{IEEEabrv,main}

\end{document}